\documentclass{ifacconf}

\usepackage{graphicx}      
\usepackage{natbib}        
\usepackage{subcaption}
\usepackage{url}

\usepackage{fancyvrb}
\usepackage{amssymb}
\usepackage{array}
\usepackage{amsmath}
\usepackage{float}
\usepackage{multirow}
\usepackage{tabularx}
\usepackage{longtable}
\usepackage{booktabs}
\usepackage{multicol}
\usepackage{listings}
\usepackage{verbatim}
\usepackage[utf8]{inputenc} 
\usepackage[dvipsnames]{xcolor}
\usepackage{color,soul}

\DeclareRobustCommand{\hlgr}[1]{{\sethlcolor{lime}\hl{#1}}}

\DeclareMathOperator*{\argmax}{argmax}
\DeclareMathOperator*{\argmin}{argmin}

\usepackage[T1]{fontenc}
\usepackage{overpic}
\begin{document}
\begin{frontmatter}

\title{A Survey on Collaborative SLAM with 3D Gaussian Splatting}

\thanks[equaCont]{Equal Contribution}

\thanks[footnoteinfo]{This work was supported by the JST SPRING, Japan
Grant Number JPMJSP2102}

\author[First]{Phuc Nguyen Xuan\thanksref{equaCont}}, 
\author[Second]{Thanh Nguyen Canh\thanksref{equaCont}}, 
\author[First]{Huu-Hung Nguyen},
\author[Second]{Nak Young Chong} and
\author[Third]{Xiem HoangVan}

\address[First]{Institute of System Integration, Le Quy Don Technical University, Hanoi, 10000, Vietnam. (e-mail:\{phucnx, hungnh\}.isi@lqdtu.edu.vn).}
\address[Second]{Graduate School of Advanced Science and Technology, Japan Advanced Institute of Science and Technology, Nomi, 923-1211, Japan. (e-mail: \{thanhnc, nakyoung\}@jaist.ac.jp)}
\address[Third]{University of Engineering and Technology, Vietnam National University, Hanoi, 10000, Vietnam. (e-mail: xiemhoang@vnu.edu.vn)}

\begin{abstract} 
This survey comprehensively reviews the evolving field of multi-robot collaborative Simultaneous Localization and Mapping (SLAM) using 3D Gaussian Splatting (3DGS). As an explicit scene representation, 3DGS has enabled unprecedented real-time, high-fidelity rendering, ideal for robotics. However, its use in multi-robot systems introduces significant challenges in maintaining global consistency, managing communication, and fusing data from heterogeneous sources. We systematically categorize approaches by their architecture-centralized, distributed-and analyze core components like multi-agent consistency and alignment, communication-efficient, Gaussian representation, semantic distillation, fusion and pose optimization, and real-time scalability. In addition, a summary of critical datasets and evaluation metrics is provided to contextualize performance. Finally, we identify key open challenges and chart future research directions, including lifelong mapping, semantic association and mapping, multi-model for robustness, and bridging the Sim2Real gap.

 
\end{abstract}


\end{frontmatter}

\section{Introduction}
Simultaneous Localization and Mapping (SLAM) is a fundamental capability in robotics and computer vision, enabling an autonomous agent to construct a map of an unknown environment while simultaneously tracking its own position within it~(\cite{kazerouni2022survey}. For applications ranging from autonomous driving~(\cite{wang2025depth}) to augmented reality~(\cite{sheng2024review}), a reliable understanding of the environment is crucial for safe navigation and intelligent decision-making. Historically, visual SLAM systems~(\cite{canh2024s3m, canh2023object}) relied on sparse representations such as point clouds, surfels, or voxels. While these methods achieved accurate and often real-time tracking, their sparse nature limited their utility for tasks requiring a detailed understanding of scene geometry and appearance, such as realistic rendering or collision avoidance. 

The field then shifted toward dense mapping, with pioneering systems like DTAM~(\cite{newcombe2011dtam}) and KinectFusion~(\cite{newcombe2011kinectfusion}) generating detailed surface models. These approaches, however, often faced challenges with high memory consumption and slow processing speeds, limiting their scalability. A significant breakthrough arrived with the advent of neural implicit representations, particularly Neural Radiance Fields (NeRFs)~(\cite{mildenhall2021nerf}), which represent a scene as a continuous function learned by a neural network. NeRF-based SLAM systems~(\cite{tosi2024nerfs, sandstrom2023point}) demonstrated an unprecedented ability to generate high-fidelity, photorealistic reconstructions and perform novel view synthesis (NVS), enabling highly detailed and immersive virtual exploration of captured scenes. Despite their quality, NeRFs suffer from significant drawbacks for real-time robotics: their reliance on large neural networks leads to slow training and rendering speeds, and their implicit nature makes it difficult to efficiently edit or update the map, a critical function for dynamic environments.

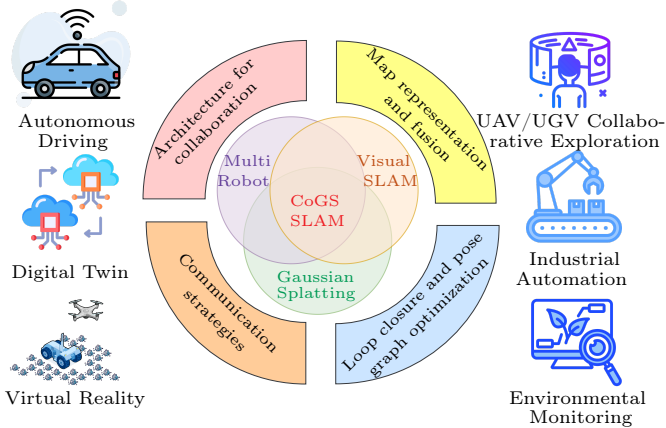
\begin{figure}
    \centering
    \begin{overpic}[width=0.48\textwidth, unit=1pt]{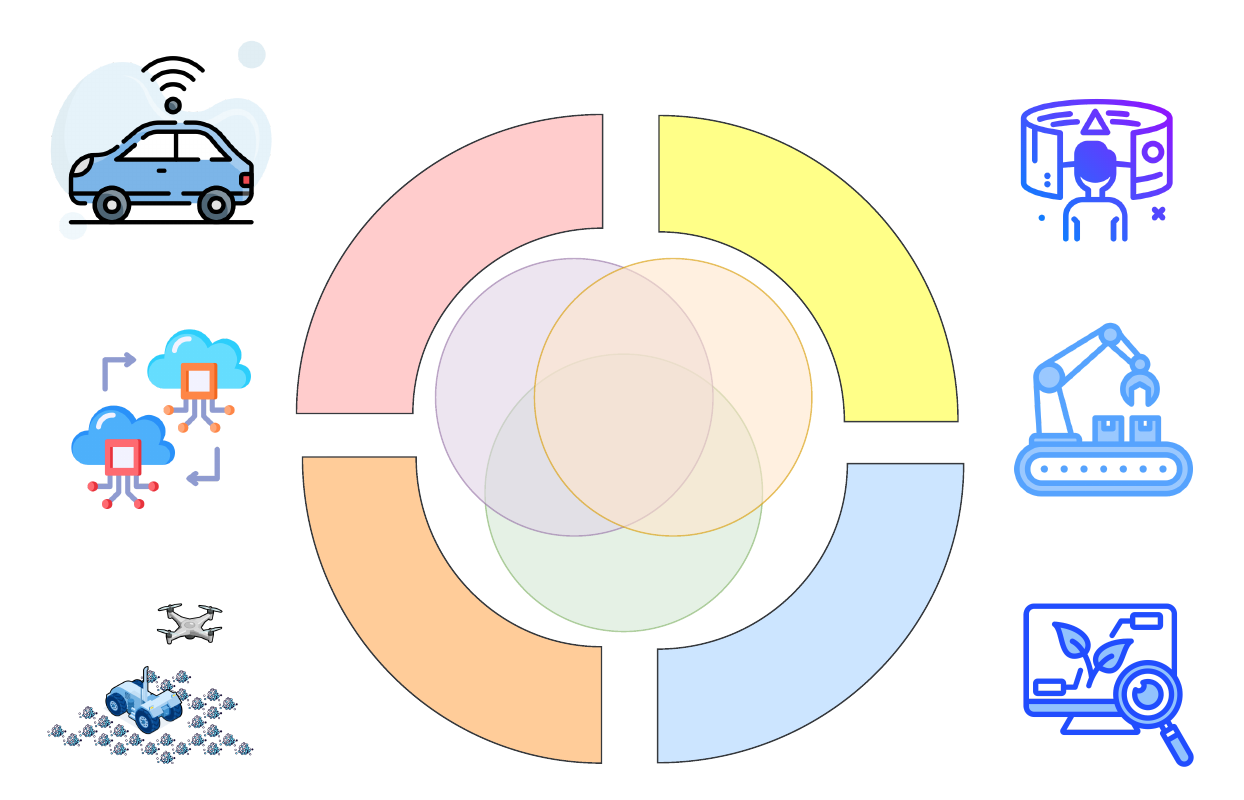}       
        \put(5.0, 42.0){\scriptsize Autonomous}
        \put(8.0, 39.0){\scriptsize Driving}
        \put(4.0, 19.0){\scriptsize Digital Twin}
        \put(74.0, 42.0){\scriptsize UAV/UGV Collabo-}
        \put(76.0, 39.0){\scriptsize rative Exploration}
        \put(3.0, 0.0){\scriptsize Virtual Reality}
        \put(82.0, 21.0){\scriptsize Industrial}
        \put(80.0, 18.0){\scriptsize Automation}
        \put(79.0, 0.0){\scriptsize Environmental}
        \put(82.0, -3.0){\scriptsize Monitoring}
        \put(44.0, 18.5){\tiny \textcolor{ForestGreen}{Gaussian}}
        \put(44.0, 16.0){\tiny \textcolor{ForestGreen}{Splatting}}

        \put(36.0, 36.0){\tiny \textcolor{RoyalPurple}{Multi}}
        \put(35.0, 33.0){\tiny \textcolor{RoyalPurple}{Robot}}

        \put(56.0, 36.0){\tiny \textcolor{Bittersweet}{Visual}}
        \put(57.0, 33.0){\tiny \textcolor{Bittersweet}{SLAM}}

        \put(46.0, 30.0){\tiny \textcolor{Red}{CoGS}}
        \put(46.0, 27.0){\tiny \textcolor{Red}{SLAM}}

        \put(25.0, 37.0){\rotatebox{45}{\tiny Architecture for}}
        \put(28.0, 36.0){\rotatebox{45}{\tiny collaboration}}

        \put(29.0, 22.0){\rotatebox{-45}{\tiny Communication}}
        \put(29.5, 16.5){\rotatebox{-45}{\tiny strategies}}

        \put(57.0, 53.0){\rotatebox{-45}{\tiny Map representation}}
        \put(60.0, 46.0){\rotatebox{-45}{\tiny and fusion}}

        \put(53.6, 4.8){\rotatebox{45}{\tiny Loop closure and pose}}
        \put(57.4, 4.0){\rotatebox{45}{\tiny graph optimization}}
        
    \end{overpic}
    \caption{Illustration of the relationship between the Collaborative Gaussian Splatting SLAM and related fields.}
    \label{fig:multirobotGS}
\end{figure}

While single-agent SLAM has seen remarkable progress, the task of mapping large-scale environments quickly and robustly remains an open problem. For applications like autonomous driving, transitioning from structured highways to complex urban scenes introduces significant challenges where traditional single-agent localization can be inadequate~(\cite{chang2021kimera, karrer2018cvi}). Employing a team of robots for collaborative SLAM (CSLAM) is essential to overcome these limitations and offers several distinct advantages~(\cite{huang2021disco}). This collaborative approach allows for accelerated, large-scale mapping, providing more comprehensive shared spatial awareness and timely situational awareness over large areas~(\cite{wang2025depth}). Furthermore, collaboration increases the robustness of the system by allowing robots to share information amongst themselves. By fusing data, the team can build a more complete and accurate map, enhancing localization accuracy where a single agent might fail. Ultimately, CoSLAM enables tasks that would be impossible for a single robot, providing a more comprehensive shared spatial awareness or ``situational awareness'' over large environments. At the same time, these multi-robot scenarios introduce their own significant challenges, such as dealing with network characteristics like time delays and bandwidth, and ensuring consistent information access and data consistency among all agents. 
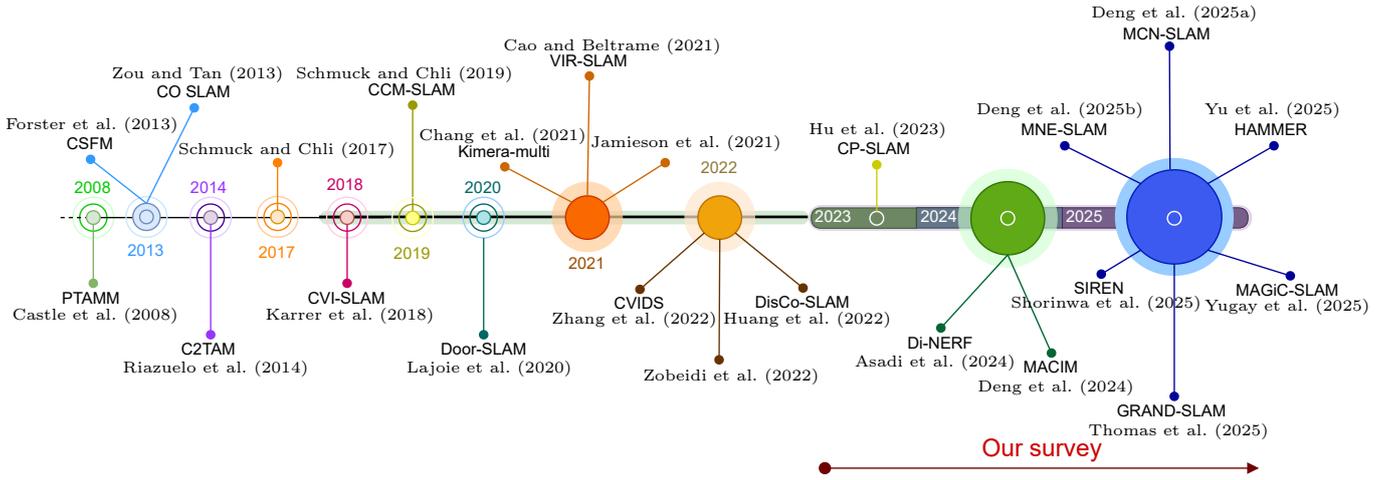
\begin{figure*}
    \centering
    \begin{overpic}[width=\textwidth, unit=1pt]{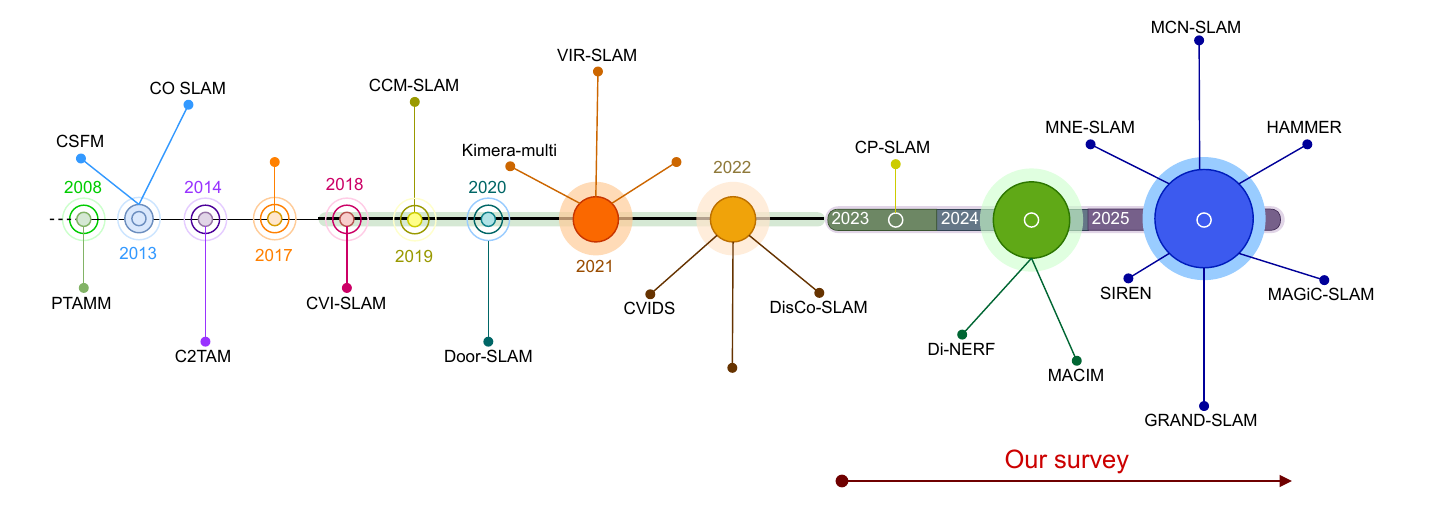}       
        \put(0.0, 13.0){\tiny \cite{castle2008video}}
        \put(-0.5, 26.9){\tiny \cite{forster2013collaborative}}
        \put(7.2, 30.6){\tiny \cite{zou2012coslam}}
        \put(8.0, 9.2){\tiny \cite{riazuelo2014c2tam}}
        \put(12.0, 25.1){\tiny \cite{schmuck2017multi}}
        \put(18.3, 13.0){\tiny \cite{karrer2018cvi}}
        \put(20.5, 30.6){\tiny \cite{schmuck2019ccm}}
        \put(28.5, 9.2){\tiny \cite{lajoie2020door}}
        \put(29.4, 26.1){\tiny \cite{chang2021kimera}}
        \put(35.5, 32.6){\tiny \cite{cao2021vir}}
        \put(41.8, 25.6){\tiny \cite{jamieson2021multi}}
        \put(39.0, 12.7){\tiny \cite{zhang2022cvids}}
        \put(45.6, 8.6){\tiny \cite{zobeidi2022dense}}
        \put(51.4, 12.7){\tiny \cite{huang2021disco}}
        \put(57.6, 26.5){\tiny \cite{hu2023cp}}
        \put(60.9, 9.6){\tiny \cite{asadi2024di}}
        \put(69.8, 7.8){\tiny \cite{deng2024macim}}
        \put(69.7, 28.0){\tiny \cite{deng2025mne}}
        \put(72.2, 13.9){\tiny \cite{shorinwa2025siren}}
        \put(78.0, 35.0){\tiny \cite{deng2025mcn}}        
        \put(77.8, 4.6){\tiny \cite{thomas2025grand}}
        \put(86.2, 28.0){\tiny \cite{yu2025hammer}}        
        \put(86.2, 13.7){\tiny \cite{yugay2025magic}}
        
    \end{overpic}
    \caption{Timeline of some noteworthy collaborative SLAM methods.}
    \label{fig:collabslam}
\end{figure*}

To address the limitations of NeRFs for robotics, 3D Gaussian Splatting (3DGS) has emerged as a revolutionary approach. Unlike implicit representations that require querying a neural network for every point in space, 3DGS is an explicit representation that models a scene as a collection of 3D Gaussian primitives, each defined by a position (mean), covariance, color, and opacity~(\cite{tosi2024nerfs}). This explicit structure provides several key advantages that make it highly suitable for robotics applications. A primary benefit is its real-time performance, as its differentiable rasterization pipeline overcomes the slow training and rendering speeds associated with large neural networks that are bottlenecks for NeRF-based systems~(\cite{naumann2024nerf}). It allows the map to be easily edited or deformed, which is beneficial for map correction and adaptation. Furthermore, 3DGS enables high-fidelity reconstruction, capturing photorealistic detail and expressive visual, geometric, and semantic information, enabling expressive scene reconstruction comparable in quality to NeRFs~(\cite{charatan2024pixelsplat, wu2024recent}). Crucially, its explicit and editable nature means the map can be directly manipulated by applying rigid-body transformations to the Gaussians. This is a vital benefit for robotics, as it simplifies essential SLAM back-end tasks like map correction after loop closure and merging submaps from different agents in collaborative settings. When paired with a visual-inertial sensor suite, 3DGS systems can also achieve metric scale estimation and gravity alignment, which is necessary for autonomous exploration and guarantees higher accuracy.

Fig~\ref{fig:multirobotGS} shows the new frontier at the intersection of these fields: Multi-robot, Visual SLAM, and 3DGS. This combination presents a unique opportunity to create large-scale, high-fidelity digital twins of the environment with unprecedented speed and detail. However, integrating these technologies also introduces a distinct set of significant research challenges:
\begin{itemize}
    \item Map Consistency and Fusion: How can multiple, independently generated Gaussian submaps be effectively merged into a single, globally consistent map without introducing artifacts or geometric distortions?
    \item Communication Bottlenecks: Centralized architectures, where all data is sent to a single server, can strain network bandwidth. Distributed systems require carefully designed protocols to share map information—be it raw data, compressed Gaussians, or model parameters—efficiently.
    \item Scalability and Robustness: Extending systems beyond controlled, small-scale indoor settings to large-scale, real-world outdoor environments or handling data from heterogeneous robot teams with different sensing modalities, challenging lighting, and reflections remains a major hurdle.
    \item Global Consistency: The classic SLAM problem of accumulated trajectory drift is amplified in multi-agent settings, making robust inter-and intra-robot loop closure mechanisms critical for maintaining long-term global consistency. Robust inter- and intra-robot loop closure mechanisms are therefore critical for maintaining long-term accuracy.
\end{itemize}

To provide a structured and comprehensive understanding of this rapidly evolving domain, this survey focuses specifically on SLAM systems that utilize 3DGS as the primary scene representation within a multi-robot collaborative framework. The overall scope of this review is visually summarized in Fig~\ref{fig:collabslam}. Our primary contributions are to:
\begin{enumerate}
    \item Propose a clear taxonomy of current systems based on their underlying system architecture (\textit{e.g.}, centralized, distributed), reflecting the primary design choices found in the literature.
    \item Synthesize the state-of-the-art by dissecting key technical components and methodologies, including multi-agent consistency, communication efficiency, Gaussian representation, fusion, and optimization.
    \item Summarize the benchmark datasets and evaluation metrics that are essential for validating and comparing the performance of different CoGS-SLAM systems.
    \item Identify critical open challenges and outline promising future research directions to guide further innovation in the field.
\end{enumerate}
The remainder of this paper is organized as follows:
Section~\ref{sec:foundation} covers the foundational principles of 3DGS and its use in single-robot SLAM. Section~\ref{sec:gscollab} then delves into the core of the survey, offering a deep dive into the key technical methodologies for collaborative 3DGS systems, including architecture, map fusion, communication, and optimization. Building on this, Section~\ref{sec:analysis} presents a comparative analysis of existing methods and their applications. Finally, Section~\ref{sec:open} discusses critical open challenges and future research directions, followed by a conclusion in Section~\ref{sec:conclusion}.


\section{Background and Foundations} \label{sec:foundation}
This section covers the foundation principles necessary to understand the main contributions of this survey. We begin by reviewing the architectures and core concepts of CoSLAM, followed by a technical overview of the 3D Gaussian Splatting representation and its application in single-robot SLAM systems.

\subsection{Collaborative SLAM}
CoSLAM extends the single-agent problem to a team of robots, with the goal of building a single, globally consistent map by sharing information among agents. This approach is crucial for efficiently mapping large-scale environments and increasing the robustness of the overall estimation process. From a probabilistic perspective, the CoSLAM problem can be formulated as a Maximum a Posteriori (MAP) estimation problem. The goal is to find a set of all robot trajectories $\mathcal{X} = \{\mathbf{x}_1, \cdots, \mathbf{x}_N \}$ for $N$ robots and the global map $\mathcal{M}$ that maximize the posterior probability, given all sensor measurements $\mathcal{Z}$ and odometry/control inputs $\mathcal{U}$:
\begin{equation}
    (\mathcal{X}^*, \mathcal{M}^*) = \argmax_{\mathcal{X}, \mathcal{M}} p(\mathcal{X}, \mathcal{M} | \mathcal{Z}, \mathcal{U}).
\end{equation}

In practice, this is typically solved using Pose Graph Optimization (PGO), which focuses on estimating trajectories of the robots. The problem is converted into a non-linear least squares minimization over a graph, where nodes represent robot poses (or keyframes) and edges represent spatial constraints between them. The objective function to be minimized is the sum of squared Mahalanobis distances of the error terms from all constraints:
\begin{equation}
\begin{aligned}
    \mathcal{X^*} = \argmin_{\mathcal{X}} \Bigg ( &\sum_{(i,j) \in \mathcal{O}} ||E_o(\mathbf{x}_i, \mathbf{x}_j, \mathbf{z}_{ij})||^2_{\sum_o} \\
    &+ \sum_{(p, q)\in \mathcal{L}} || E_l (\mathbf{x}_p, \mathbf{x}_q, \mathbf{z}_{pq})||^2_{\sum_l} \Bigg )
\end{aligned},
\end{equation}
where $E_o$ is the error from odometric constraints $\mathcal{O}$ between consecutive poses, and $E_l$ is the error from loop closure constraints $\mathcal{L}$, which are non-sequential constraints that are critical for correcting accumulated drift. In distributed systems, this global optimization is solved iteratively via distributed PGO (dPGO)~(\cite{li2024distributed}), where agents solve parts of the problem locally and communicate updates to their neighbors.

The CoSLAM systems are typically categorized by their communication architecture, which includes centralized or distributed~(\cite{zou2019collaborative}). In this client-server model, agents stream local data (\textit{e.g.}, keyframes, sensor measurements) to a powerful central server. The server handles computationally intensive tasks like map fusion, global optimization, and multi-agent loop closure detection~(\cite{zhang2022cvids, liu2025cpl}). Some systems use two-way communication, where the server sends optimized results back to the agents to refine their local estimates. On the other hand, in a distributed paradigm, robots communicate directly with their neighbors in a peer-to-peer fashion to exchange map information, detect inter-robot loop closures, and perform distributed dPGO~(\cite{lajoie20253d}). This approach is generally more scalable and robust to single points of failure, but makes achieving global consistency more challenging.

In addition, the key to creating a globally consistent map is robust data association, which involves finding both intra-robot loop closures (a robot re-observing its own past trajectory) and inter-robot loop closures (a robot observing an area previously mapped by another agent). This is typically achieved using place recognition techniques that compare descriptors of keyframes, such as Bag-of-Words (BoW), NetVLAB, or Scan Context for LiDAR. A major perceptual aliasing issue can generate false-positive loop closures. To address this, robust back-ends incorporate outlier rejection mechanisms, such as Pairwise Consistent Measurement (PCM) set maximization or Graduate Non-Convexity (GNC)~(\cite{chang2021kimera, tian2022kimera}). Another core module is map fusion, which merges local maps or submaps from different agents into a single, coherent representation. This process is typically triggered after a successful inter-robot loop closure provides the relative transformation, $T_{ij} \in SE(3)$, between the coordinate frames of robot $i$ and robot $j$. For explicit map representations (\textit{e.g.}, point clouds, meshes, 3D Gaussians), the data from one map is transformed into the reference frame of the other and then merged geometrically. 

\subsection{Foundations of 3D Gaussian Splatting (3DGS)}

3D Gaussian Splatting is an explicit scene representation that models a scene as a collection of 3D Gaussian primitives, each representing a part of the scene's geometry and appearance. Unlike implicit representations like NeRFs, this explicit structure is more amenable to traditional SLAM back-ends and allows for direct manipulation of the scene geometry~(\cite{zhu20243d}). The fundamental building block of 3DGS is an anisotropic 3D Gaussian, whose density is described by:
\begin{equation}
    G(\mathbf{x}) = e^{-\frac{1}{2}(\mathbf{x}-\mu)^T \sum^{-1} (\mathbf{x-\mu})},
\end{equation}
where $\mathbf{x}$ is a point in 3D space, $\mu$ is the center of the Gaussian, and $\sum$ is the covariance matrix. Each Gaussian is defined by the set of optimizable parameters: position $\mu \in \mathbb{R}^3$, covariance $\sum \in \mathbb{R}^{3 \times 3}$ decomposed for optimization into a rotation quaternion $\mathbf{q} \in \mathbb{R}^4$ and a scaling vector $\mathbf{s} \in \mathbb{R}^3$, color $c$, and opacity $\alpha \in [0,1]$. The core of 3DGS is its differentiable rasterization pipeline, which allows gradients to be backpropagated from a 2D image loss to the 3D Gaussian parameters~(\cite{wu2024recent}). For a given camera viewpoint, the 3D Gaussians are projected onto the 2D image plane. The 3D covariance matrix $\sum$ is projected into a 2D covariance matrix $\sum'$ using the Jacobian $\mathbf{J}$ of the perspective projection $\sum' = \mathbf{J} \sum \mathbf{J}^T$. For efficiency, the rendering process uses a tile-based rasterizer. The screen is split into tiles, and Gaussians are sorted by depth within each tile, enabling massive parallelization. The final color $c$ of a pixel is then computed by alpha-blending the $N$ Gaussians that overlap it, ordered by depth:
\begin{equation}
    c = \sum_{i=1}^N c_i \alpha_i' \prod_{j=1}^{i-1}(1-\alpha_j'),
\end{equation}
where $c_i$ is the color of the $i-th$ Gaussian and $\alpha_j'$ is its opacity modulated by the 2D Gaussian evaluation at the pixel center~(\cite{bao20253d}).

\subsection{Gaussian Splatting in Single-Robot SLAM}
The principles of 3DGS have been successfully integrated into single-robot SLAM frameworks, leading to a new class of systems capable of real-time, high-fidelity dense mapping. Pioneering works in this area, such as SplaTAM~(\cite{keetha2024splatam}), GS-SLAM~(\cite{yan2024gs}), and LoopSplat~(\cite{zhu2025loopsplat}), have demonstrated the effectiveness of using 3DGS as the core map representation. These systems typically follow the classic SLAM paradigm with specialized modules for tracking and mapping. The tracking module estimates the camera's 6-DoF pose for each incoming frame. This is commonly formulated as an optimization problem that minimizes a loss function composed of photometric and geometric errors between the rendered images ($\hat{C}$, $\hat{D}$) and the observed geometric-truth images ($C_{gt}, D_{gt}$). The tracking loss $\mathcal{L}_{track}$ is typically a weighted sum $\lambda_{depth}$ of a color loss and a depth loss:
\begin{equation}
    \mathcal{L}_{track} = \mathcal{L}_{color} + \lambda_{depth}\mathcal{L}_{depth},
\end{equation}
where $\mathcal{L}_{color}$ can be an L1 loss, potentially combined with a structural similarity (SSIM) term, and $\mathcal{L}_{depth}$ is often an L1 loss on the depth maps. Because the rendering pipeline is differentiable, gradients from this loss are backpropagated to the camera pose parameters, which are then updated via gradient descent. 

The mapping module is responsible for building and refining the 3DGS map using selected keyframes. New Gaussians are first seeded using points from the new keyframe's depth map. Then, the parameters of the Gaussians and the poses of a local window of keyframes are optimized by minimizing a mapping loss, $\mathcal{L}_{map}$. This loss is similar to the tracking loss, but may include an additional regularization term $\mathcal{L}_{reg}$, weighted by $\lambda_{reg}$, to maintain good geometric properties for the Gaussians:
\begin{equation}
    \mathcal{L}_{map} = \mathcal{L}_{color} + \lambda_{depth}\mathcal{L}_{depth} + \lambda_{reg}\mathcal{L}_{reg}.
\end{equation}
The adaptive density control mechanisms-densification and pruning-are also applied during this stage to dynamically refine the map's geometry. To combat accumulated trajectory drift, many GS-SLAM systems incorporate loop closure. When a loop is detected, a PGO is performed over the keyframe trajectory. The PGO minimizes an objective function that includes both odometric constraints ($\varepsilon_{odom}$) and loop closure constraints ($\varepsilon_{loop}$):
\begin{equation}
    E(\mathbf{T}) = \sum_{(i, j) \in \varepsilon_{odom}} e_{ij}^T\mathbf{\Omega}_{ij}e_{ij} + \sum_{(p,q) \in \varepsilon_{loop}} e_{ij}^T\mathbf{\Omega}_{pq}e_{pq},
\end{equation}
where $\mathbf{T}$ is the set of all keyframe poses, $\mathbf{\Omega}$ is the information matrix for a constraint, and the error term $e_{ij}$ is often defined in the tangent space $\mathfrak{se}(3)$ as $e_{ij} = \log((\bar{T}_{ij})^{-1}(T_i^{-1}T_j))$.

\section{Collaborative SLAM with Gaussian Splatting} \label{sec:gscollab}

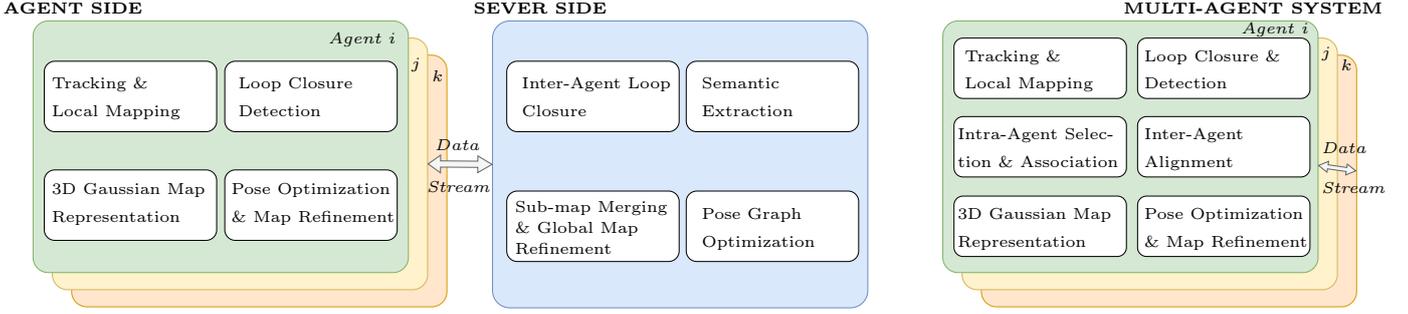
\begin{figure*}[!ht]
    \centering
    \begin{overpic}[width=\textwidth, unit=0.5pt]{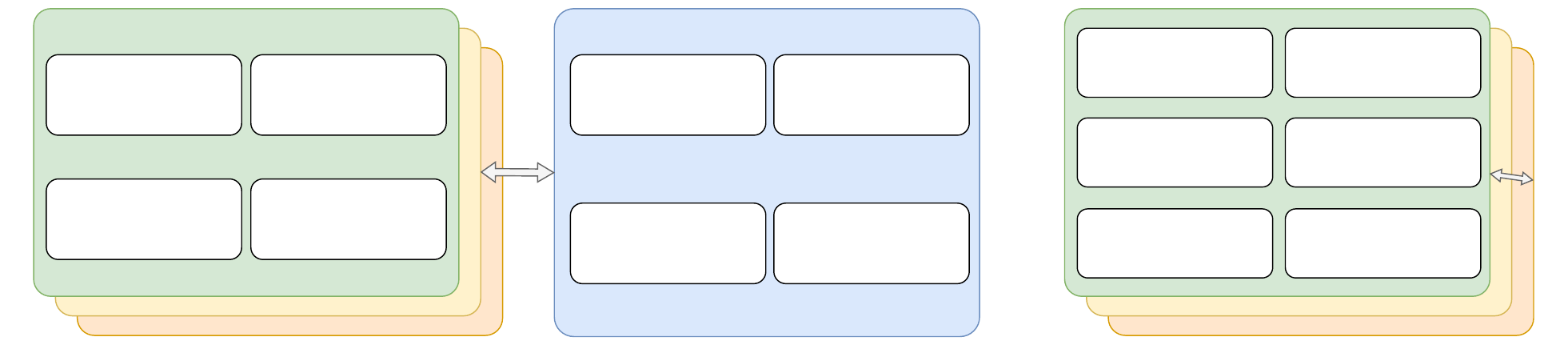}       
        \put(0.0, 22.0){\tiny \textbf{AGENT SIDE}} 
        
        \put(23.5, 19.8){\tiny \textit{Agent $i$}} 
        \put(29.5, 18.0){\tiny \textit{$j$}} 
        \put(31.0, 17.0){\tiny \textit{$k$}}
        
            \put(3.5, 16.5){\tiny Tracking \&}
            \put(3.5, 14.5){\tiny Local Mapping}

            \put(17.0, 16.5){\tiny Loop Closure}
            \put(17.0, 14.5){\tiny Detection}
         
            \put(3.5, 8.8){\tiny 3D Gaussian Map}
            \put(3.5, 6.8){\tiny Representation}

            \put(16.5, 8.8){\tiny Pose Optimization }
            \put(16.5, 6.8){\tiny \& Map Refinement}

        \put(31.2, 12.0){\tiny \textit{Data}}
        \put(30.6, 9.0){\tiny \textit{Stream}}
        
        \put(34.0, 22.0){\tiny \textbf{SEVER SIDE}} 
            \put(37.5, 16.5){\tiny Inter-Agent Loop }
            \put(37.5, 14.5){\tiny Closure}

            \put(50.5, 16.5){\tiny Semantic }
            \put(50.5, 14.5){\tiny Extraction}

            \put(37.0, 7.5){\tiny Sub-map Merging }
            \put(37.0, 6.0){\tiny \& Global Map}
            \put(37.0, 4.5){\tiny Refinement}

            \put(50.5, 7.0){\tiny Pose Graph }
            \put(50.5, 5.0){\tiny Optimization}
        
        \put(81.0, 22.0){\tiny \textbf{MULTI-AGENT SYSTEM}} 
        \put(89.5, 20.5){\tiny \textit{Agent $i$}} 
        \put(95.3, 18.8){\tiny \textit{$j$}} 
        \put(96.7, 17.8){\tiny \textit{$k$}}
        \put(95.3, 11.8){\tiny \textit{Data}}
        \put(95.3, 8.9){\tiny \textit{Stream}}
        
            \put(69.5, 18.5){\tiny Tracking \& }
            \put(69.5, 16.5){\tiny Local Mapping}

            \put(82.5, 18.5){\tiny Loop Closure \& }
            \put(82.5, 16.5){\tiny Detection}

            \put(69.0, 12.8){\tiny Intra-Agent Selec- }
            \put(69.0, 10.8){\tiny tion \& Association}

            \put(82.5, 12.8){\tiny Inter-Agent}
            \put(82.5, 10.8){\tiny Alignment}

            \put(69.0, 7.0){\tiny 3D Gaussian Map }
            \put(69.0, 5.0){\tiny Representation}
            
            \put(82.5, 7.0){\tiny Pose Optimization}
            \put(82.5, 5.0){\tiny \& Map Refinement}
            
    \end{overpic}
    \caption{Core Components of CoGS-SLAM: Centralized (left), Distributed (right).}
    \label{fig:Arch}
\end{figure*}

This section forms the core of our survey. We begin by providing a formal problem statement for collaborative Gaussian Splatting SLAM (CoGS-SLAM). We then delve into the unique challenges this paradigm presents and review the representative architectures and cross-cutting design patterns that have emerged in the literature to address them.
\subsection{Problem Formulation}

We consider a team of $N$ robots $\mathcal{R} = \{R_1, \dots, R_N\}$, collaboratively exploring an unknown 3D environment. Each robot $R_i$ is equipped with sensors (\textit{e.g.}, RGB-D camera, LiDAR), and its task is to incrementally build a scene representation based on Gaussian splatting while maintaining globally consistent poses. It provides a time-ordered sequence of measurements:
\begin{equation}
\mathcal{D}_i = \{(I_t, \mathbf{z}_t)\}_{t=1}^{T_i},
\end{equation}
where $I_t$ denotes an RGB frame and $\mathbf{z}_t$ may include depth, or inertial readings. From this data, each robot constructs a \emph{local Gaussian submap}, represented as a set of 3D Gaussians:
\begin{equation}
\mathcal{M}_i^{\text{local}} = 
\left\{
\mathbf{g}_k = \big(\boldsymbol{\mu}_k, \mathbf{\Sigma}_k, \mathbf{a}_k^{\text{SH}}, \alpha_k, \mathbf{f}_k\big)
\right\}_{k=1}^{K_i},
\end{equation}
where:
\begin{itemize}
    \item $\boldsymbol{\mu}_k \in \mathbb{R}^3$ is the Gaussian center,
    \item $\mathbf{\Sigma}_k \in \mathbb{R}^{3 \times 3}$ is the covariance,
    \item $\mathbf{a}_k^{\text{SH}}$ are spherical harmonic coefficients for view-dependent appearance,
    \item $\alpha_k \in [0,1]$ is the opacity,
    \item $\mathbf{f}_k \in \mathbb{R}^d$ is an optional semantic feature vector.
\end{itemize}

The primary objective is to jointly estimate two quantities:
\begin{enumerate}
    \item \textbf{\textit{Globally consistent robot trajectories}} $\{\mathbf{T}_i(t)\}$, where $\mathbf{T}_i(t) \in SE(3)$ transforms from robot $i$'s local frame to a common global reference frame.
    \item A fused \textbf{\textit{global Gaussian map}}$ 
        \mathcal{M}^{\text{global}} = \bigcup_{i=1}^N \mathbf{T}_i^\star \cdot \mathcal{M}_i^{\text{local}}$, which is the union of all properly transformed local submaps, spatially aligned and semantically consistent.
\end{enumerate}

This is typically formulated as a joint optimization problem that minimizes a combination of loss terms over robot poses and Gaussian parameters:
\begin{equation}
\begin{split}
\min_{\{\mathbf{T}_i(t)\}, \{\mathcal{M}_i^{\text{local}}\}} 
\lambda_{\mathrm{track}} \,\mathcal{L}_{\text{track}} +
\lambda_{\mathrm{map}} \,\mathcal{L}_{\text{map}} + \\ \lambda_{\mathrm{loop}} \,\mathcal{L}_{\text{loop}} +
\lambda_{\mathrm{fuse}} \,\mathcal{L}_{\text{fuse}},
\end{split}
\label{eq:joint_opt}
\end{equation}
where:
\begin{align}
\mathcal{L}_{\text{track}} &= \sum_{i=1}^N \sum_{t} 
\left\| I_t - \Pi\big(\mathbf{T}_i(t) \cdot \mathcal{M}_i^{\text{local}}\big) \right\|_1, \\ 
\mathcal{L}_{\text{map}} &= \sum_{i=1}^N \sum_{t,k} 
w_{k,t} \cdot \left\| I_t - \hat{I}_t(\mathbf{g}_k) \right\|_1, \\
\mathcal{L}_{\text{loop}} &= \sum_{(p,q) \in \mathcal{E}_{\text{LC}}} 
\rho\!\left(\big\|\log(\mathbf{T}_p^{-1} \mathbf{T}_q) - \hat{\boldsymbol{\xi}}_{pq} \big\|_{\Sigma_{pq}^{-1}}^2\right), \\
\mathcal{L}_{\text{fuse}} &= \sum_{(i,j) \in \mathcal{E}}
\rho\!\left(\big\|\log(\mathbf{T}_{ij}^{-1} \mathbf{T}_i^{-1} \mathbf{T}_j)\big\|_{\Sigma_{ij}^{-1}}^2\right). 
\end{align}
Here, $\lambda_\ast$ are balancing weights for each loss term.
Where $\Pi(\cdot)$ denotes a differentiable Gaussian renderer, $\rho(\cdot)$ a robust loss, $\mathcal{E}_{\text{LC}}$ loop closure constraints, and $\mathcal{E}$ the set of inter-agent alignment edges. The loss terms balance photometric consistency ($\mathcal{L}_{\text{track}}, \mathcal{L}_{\text{map}}$) with geometric alignment ($\mathcal{L}_{\text{loop}}, \mathcal{L}_{\text{fuse}}$). Since these operate in heterogeneous units (pixels vs. metric space), practical systems employ covariance normalization or empirical weights to ensure stable optimization~(\cite{shorinwa2025siren, thomas2025grand, yu2025hammer, yugay2025magic}). By jointly reasoning over trajectories and Gaussian parameters, it captures the essential requirements of collaborative mapping: robust localization, scalable data fusion, and high-fidelity scene representation suitable for robotics.






\begin{table*}[t]
\centering
\caption{Comparison of representative Collaborative Gaussian Splatting SLAM (CoGS-SLAM) and related neural implicit SLAM systems.}
\renewcommand{\arraystretch}{1.1}
\setlength{\tabcolsep}{4pt}
\begin{tabular}{lccccc}
\hline
\textbf{System} & \textbf{Representation} & \textbf{Collaboration} & \textbf{Consistency / Fusion} & \textbf{Semantic} & \textbf{Input} \\
\hline
CP-SLAM~(\cite{liu2025cpl})        & Neural Point & Centralized & Federated avg., PGO & None & RGB-D \\
MCN-SLAM~(\cite{deng2025mcn})      & Hybrid (Tri-plane + MLP) & Distributed & Online distillation & None & RGB-D \\
MNE-SLAM~(\cite{deng2025mne})      & Hybrid (Planar-Grid + MLP)  & Distributed & Online distillation & None & RGB-D \\
Di-NeRF~(\cite{asadi2024di})   & NeRF & Distributed & Consensus ADMM & None & RGB \\
GRAND-SLAM~(\cite{thomas2025grand}) & 3DGS & Centralized & PGO, coarse-to-fine loop & Implicit (DINO) & RGB-D \\
HAMMER~(\cite{yu2025hammer})       & 3DGS  & Centralized & Frame align., global opt. & CLIP-based & RGB \\
MAC-Ego3D~(\cite{xu2025mac})       & 3DGS & Distributed & Consensus & None & RGB-D \\
MAGiC-SLAM~(\cite{yugay2025magic})  & 3DGS          & Centralized & PGO      & None                  & RGB-D \\

\hline
\end{tabular}
\label{tab:gs_collab}
\end{table*}
\subsection{Architecture in CoGS-SLAM}
The collaborative architectures for Gaussian Splatting SLAM draw on diverse strategies, primarily categorized by how agents communicate and how global map consistency is maintained: distributed and centralized. While traditional collaborative SLAM has seen a trend toward distributed systems to improve scalability and robustness, the high computational cost of 3DGS training has led to the current prevalence of centralized designs in CoGS-SLAM. Fig.~\ref{fig:Arch} shows how local Gaussian-based tracking and global consensus optimization are integrated into a unified collaborative SLAM pipeline. Table~\ref {tab:gs_collab} provides a comparative summary of representative collaborative methods from the literature, including contrasts with neural implicit approaches that are frequently used as baselines against GS-based SLAM systems.

In a distributed framework, agents operate as independent nodes, performing their own mapping and tracking while sharing information directly via peer-to-peer communication. This approach avoids a single point of failure and is theoretically more scalable. Recent advances in this area have focused on neural implicit representations. Frameworks like MNE-SLAM~(\cite{deng2025mne}), Di-NeRF~(\cite{asadi2024di}), and MCN-SLAM~(\cite{deng2025mcn}) demonstrate complementary advances in balancing scalability, efficiency, and representational fidelity. MNE-SLAM prioritizes communication efficiency and pose consistency through a distributed mapping and camera-tracking pipeline, employing a two-scale parametric-coordinate encoding that mitigates the cubic memory growth of voxel-based approaches. Meanwhile, MCN-SLAM enhances representation scalability via a hybrid planar-grid-coordinate encoding and an online distillation mechanism, yielding high-fidelity reconstruction in large and unbounded environments while reducing memory and bandwidth requirements. Collectively, these systems outline the emerging design trade-offs among communication efficiency, optimization stability, and representational richness, defining the current frontier of distributed neural SLAM architectures. Extending this decentralization, Di-NeRF focuses on the optimization strategy itself, using a consensus-driven approach like Alternating Direction Method of Multipliers (ADMM). This allows multiple agents to jointly refine Neural Radiance Field parameters and relative poses, improving robustness and fault tolerance at the expense of higher computational cost and potential convergence instability:
\begin{equation}
    \mathcal{L}_i(\theta_i) = L_i(\theta_i) + \sum_{j\in \mathcal{N}_i} (y_{ij} \Big (\theta_i - r_{ij}) + \frac{\rho}{2}||\theta_i - z_{ij}||^2_2 \Big ),
\end{equation}
where $L_i(\theta_i)$ is the local loss for agent $i$ with its local model parameters $\theta_i$, and the additional terms enforce consensus with neighboring agents $j \in \mathcal{N}_i$ using dual variables $y_{ij}$ and a penalty factor $\rho$. Beyond implicit representations, MAC-Ego3D~(\cite{xu2025mac}) introduces an explicit, real-time multi-agent CoGS-SLAM framework that achieves photorealistic reconstruction through a Multi-Agent Gaussian Consensus mechanism integrating intra- and inter-agent alignment. Unlike the coordinate-encoded neural fields used in MNE-SLAM or Di-NeRF, its explicit Gaussian parameterization enables real-time differentiable rendering and scalable collaboration. While the current implementation focuses on sequential multi-agent collaboration without communication bandwidth constraints, future extensions aim to generalize the framework to parallel agents and larger-scale, multi-room environments.

In contrast, the architectural evolution of collaborative SLAM has historically followed a transition from centralized to distributed paradigms~(\cite{lajoie2020door, tian2022kimera}), driven by the need to alleviate communication and scalability bottlenecks in multi-robot settings. However, this trend exhibits a temporary inversion in GS-SLAM, where most current frameworks remain centralized due to the heavy computational and synchronization demands of neural field optimization and radiance fusion. In practice, recent CoGS-SLAM systems adopt a server-based architecture, separating agent-side modules, responsible for tracking, local mapping, and semantic feature extraction from server-side modules, which manage cross-agent loop closure, Gaussian-to-Gaussian registration, global map merging, and solving a global PGO to ensure all submaps are aligned in a consistent frame. Collectively, these components form the functional backbone of current CoGS-SLAM frameworks, enabling high-fidelity collaborative mapping despite their centralized dependency.

In these systems, each agent handles local tracking and mapping, while a central server manages loop closure, pose graph optimization, and submap fusion, ensuring global consistency. Scene representations vary from implicit neural points (CP-SLAM~(\cite{liu2025cpl}) to explicit Gaussian splats (GRAND-SLAM~(\cite{thomas2025grand}), HAMMER~(\cite{yu2025hammer}), MAGiC-SLAM~(\cite{yugay2025magic}), with some frameworks incorporating semantic or hybrid feature fields. Centralized coordination is particularly advantageous for managing dense or implicit map representations, offloading computationally intensive tasks such as 3D Gaussian training, and aligning heterogeneous data from multiple agents. While these designs support high-fidelity reconstruction and robust map integration, they limit scalability and fault tolerance, highlighting a key architectural challenge in multi-agent GS-SLAM. Consequently, centralized GS-SLAM provides a practical compromise between performance and complexity, pending future research into distributed alternatives for large-scale or heterogeneous environments.

\subsection{Sub-problems}
\begin{figure}[!ht]
    \centering
    \includegraphics[width=0.7\linewidth]{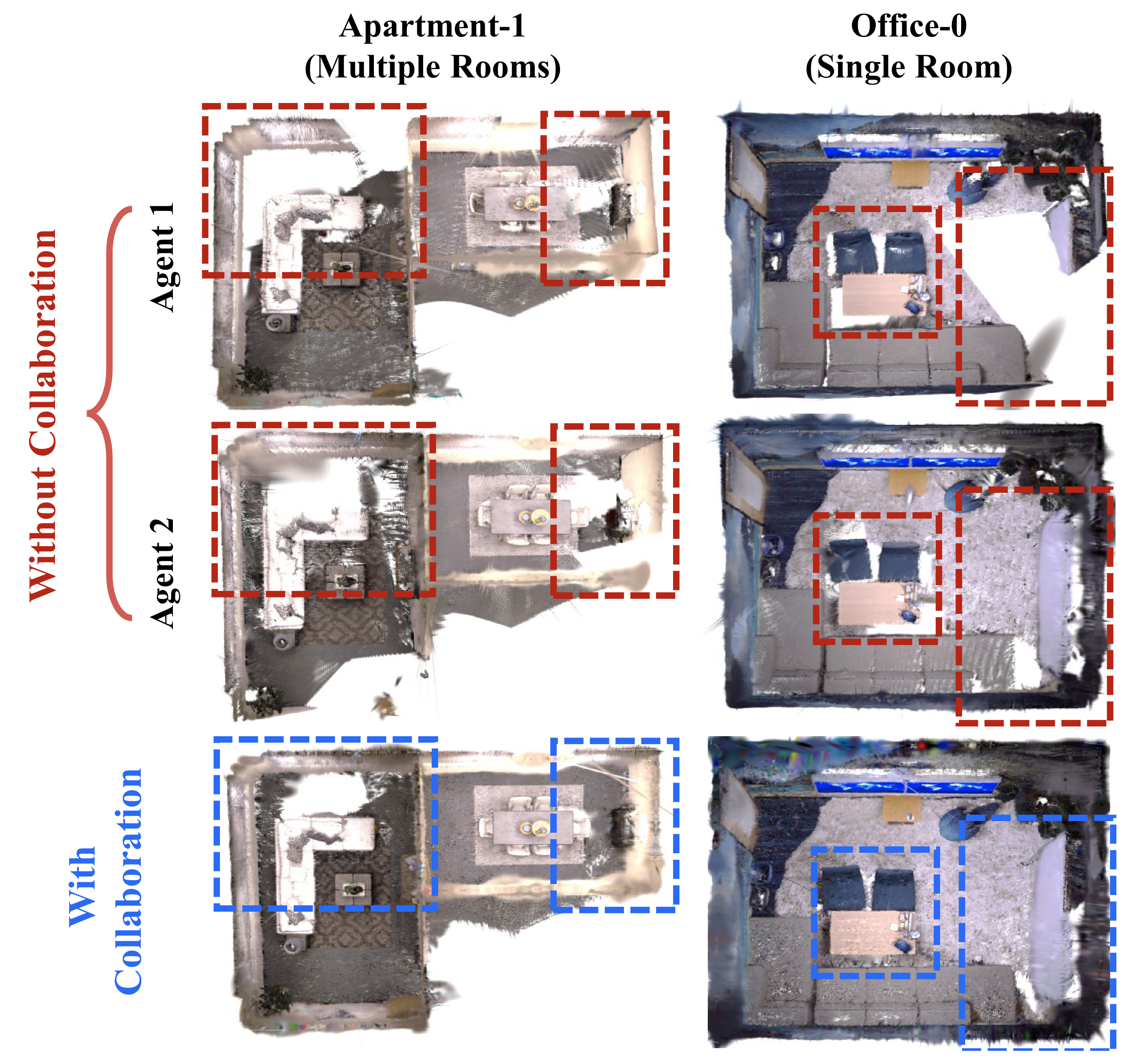}
    \caption{Compare with and without collaboration in MAC-Ego3D using Replica dataset~(\cite{straub2019replica}).}
    \label{fig:ego}
\end{figure}
CoGS-SLAM inherits classical challenges of multi-agent mapping: global consistency under drift, communication efficiency, and scalable optimization, while introducing new difficulties arising from its dense photometric representation. These include maintaining radiance-space coherence across agents and handling bandwidth-intensive Gaussian updates. Among these, we focus on two central subproblems that most critically define current CoGS-SLAM design: multi-agent consistency and alignment, and communication-efficient collaborative optimization.

\subsubsection{Multi-agent consistency and alignment:}
A central challenge in CoSLAM is achieving global consistency across the maps generated by multiple agents. In classical SLAM, this is primarily a geometric alignment problem, solved by estimating relative poses and performing a global Pose Graph Optimization~(\cite{dissanayake2001solution, kummerle2011g}). However, in CoGS-SLAM, consistency extends beyond geometric alignment to also demand radiance-space coherence across independently optimized maps. Small pose drifts can produce large photometric inconsistencies, such as duplicate surfaces, misaligned shading, and visual artifacts. This dual consistency requirement has led to various strategies, which differ based on whether the underlying map representation is explicit, implicit (neural fields), or hybrid. The foundation for consistency remains a global pose graph. Inter-agent loop closures are found by matching high-level descriptors (\textit{e.g.}, NetVLAD, DINOv2) between keyframes, providing constraints between the pose graphs of different agents. The system then seeks to find the set of globally consistent poses $\mathbf{X}^*$ by minimizing the error over all odometric and loop closure constraints. After trajectories are aligned, the Gaussian submaps must be merged. Some frameworks approach this as a joint optimization problem. MAC-Ego3D, for example, which is shown in Fig.~\ref{fig:ego} introduces a Multi-Agent Gaussian Consensus mechanism that jointly optimizes all agent poses $\mathbf{P}_i$  and a single global Gaussian map $\mathbf{G}$ by minimizing a shared photometric loss:
\begin{equation}
    \min_{\{\mathbf{P}_i\}_{i=1}^N, \mathbf{G}} \sum_{i=1}^N \sum_{j=1}^{T_i}\mathcal{L}_{photo}(\mathbf{I}_{i,j}, \mathbf{P}_{i,j}, \mathbf{G}),
\end{equation}
where $\mathcal{L}_{photo}$ is the rendering loss between the ground-truth image $\mathbf{I}_{i,j}$  and the image rendered from the global map $\mathbf{G}$ using the agent $i$'s pose $\mathbf{P}_{i,j}$. Other systems, like MAGiC-SLAM~(\cite{yugay2025magic}), perform a rigid merge followed by a non-rigid refinement of the Gaussian parameters to remove visual seams and ensure a coherent appearance. Meanwhile, HAMMER introduces global semantic anchors and a shared radiance optimization scheme to align heterogeneous agents. Nevertheless, both approaches still depend on geometric surrogates rather than direct optimization of Gaussian parameters; consequently, their alignment corrections fail to propagate photometric consistency globally, often resulting in subtle seams or discontinuities in the rendered appearance.

\begin{figure}[!ht]
    \centering
    \includegraphics[width=0.8\linewidth]{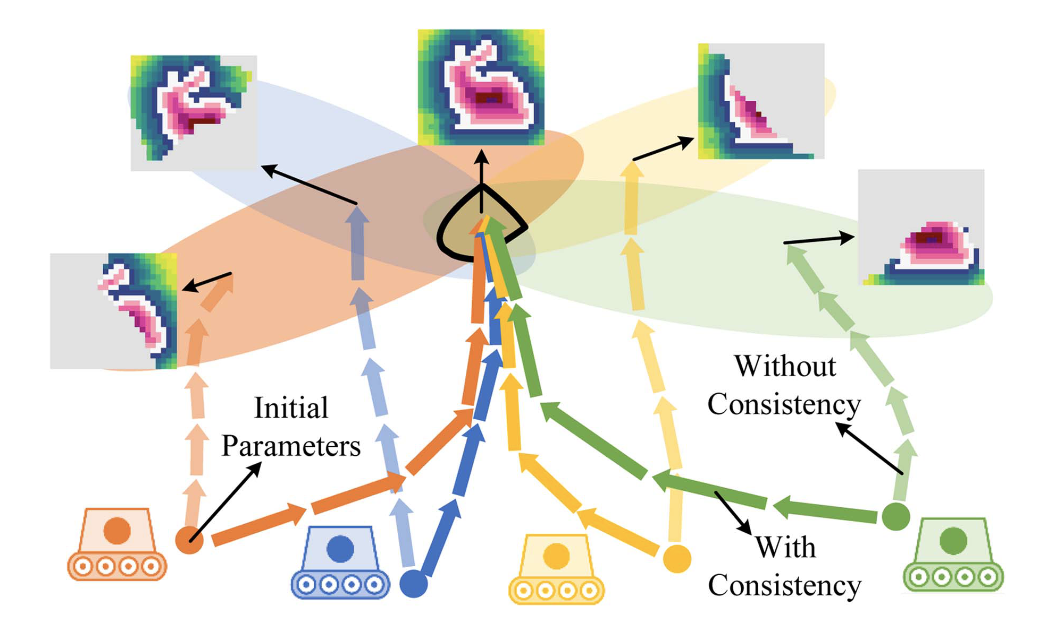}
    \caption{Illustration of collaborative consistency with and without its application.}
    \label{fig:mcn}
\end{figure}



For implicit representations like NeRFs, consistency is achieved by ensuring the underlying neural network parameters ($\theta$) are consistent across agents, as shown in Fig.~\ref{fig:mcn}, rather than by aligning explicit geometry. Systems like Di-NeRF and MACIM~(\cite{deng2024macim} formulate this as a distributed optimization problem where agents aim to find a consensus on the model parameters. Frameworks like MNR-SLAM~(\cite{deng2025mne}) and MCN-SLAM~(\cite{deng2025mcn}) use online knowledge distillation to enforce photometric consistency. When two agents observe an overlapping area, one agent's network is trained to reproduce the rendering of the other agent's network. Some systems use a hybrid approach that combines elements of both explicit and implicit representations. CP-SLAM~(\cite{hu2023cp}), for example, uses a shared map of neural points, where each point is an explicit 3D entity associated with a learnable neural feature vector. Consistency is maintained by jointly optimizing the poses of all agents and the features of the shared neural point map in a centralized back-end, minimizing a global rendering loss summed over all agent and their observations of the shared map $\mathbf{M}_{pts}$:
\begin{equation}
    \min_{\{\mathbf{T}_i\}_{t=1}^N, \mathbf{M}_{pts}} \sum_{i=1}^N \sum_t \mathcal{L}_{render} (\mathbf{I}_{i,t}, \mathbf{T}_{i,t}, \mathbf{M}_{pts}).
\end{equation}

In summary, while traditional CoSLAM achieves global consistency through geometric graph optimization, CoGS-SLAM extends this goal by introducing a dual consistency requirement that jointly enforces geometric accuracy and radiometric coherence. This shift calls for new formulations that integrate pose-graph optimization with radiance-field regularization.

\subsubsection{Communication-efficient collaborative optimization:} %
Ensuring geometric–radiometric consistency guarantees local coherence among agents; however, maintaining this level of performance in practical multi-robot environments necessitates communication-efficient collaborative optimization. This subproblem focuses on how distributed agents can exchange, update, and fuse radiance-field representations under strict bandwidth, memory, and latency constraints. Instead of transmitting dense radiance fields or entire Gaussian maps, recent systems share compact representations such as network weights, keyframe descriptors, or filtered submaps. Centralized 3DGS-based frameworks improve efficiency by transmitting only a curated subset of data. HAMMER streams posed RGB-D data for selected keyframes to a central server for offloaded 3DGS training and inter-robot alignment. GRAND-SLAM~(\cite{thomas2025grand}) extends this paradigm with incremental submap construction and rigid correction in the global frame, while emphasizing future directions in compression and communication reduction. A more granular approach is taken by MAGiC-SLAM~(\cite{yugay2025magic}), which transfers only visible splats within each agent’s current camera frustum for centralized pose correction. Collectively, these strategies balance photorealistic fidelity and scalability but remain reliant on centralized coordination, leaving distributed optimization for Gaussian representations an open challenge.

Distributed optimization avoids a central bottleneck but requires sophisticated protocols for agents to reach consensus with minimal data exchange. MAC-Ego3D introduces a multi-agent Gaussian consensus process that jointly optimizes poses and radiance parameters under an asynchronous photometric–geometric objective, maintaining real-time performance by avoiding the need to transmit full maps. SIREN~(\cite{shorinwa2025siren}) uses semantic priors to find a robust initial transformation by matching a small number of compact semantic features, which reduces the communication overhead for establishing the first collaboration link. GRAND-SLAM~(\cite{thomas2025grand}) ensures long-term stability via local submap refinement and rigid global correction. Complementary implicit-field frameworks such as Di-NeRF and MNE-SLAM~(\cite{deng2025mne}) demonstrate distributed consensus and online distillation to maintain cross-agent coherence without tight synchronization. Together, these developments highlight an emerging trend toward unified, communication-efficient optimization, where geometric consistency, radiometric coherence, and scalability are co-optimized across collaborating agents.

\subsection{Design patterns and cross-cutting strategies}
Despite architectural differences in their focus on semantic reasoning, geometric consistency, or real-time consensus, recent CoGS-SLAM frameworks reveal a set of recurring architectural regularities. These design patterns offer a coherent lens through which to understand how these systems achieve scalable, accurate, and communication-efficient mapping. Across these frameworks, several cross-cutting strategies repeatedly emerge: linking representation, optimization, and communication layers into unified pipelines. Specifically, four recurring design paradigms can be identified:
(i) shared Gaussian representations and semantic distillation,
(ii) inter-agent correspondence and consistency mechanisms,
(iii) asynchronous fusion and pose optimization, and
(iv) compression-oriented efficiency for real-time scalability.

\subsubsection{Shared Gaussian representations and semantic distillation:} Recent CoGS-SLAM systems are converging toward shared Gaussian representations as a unifying architectural motif. These frameworks encode the environment through compact Gaussian primitives that capture geometry, appearance, and uncertainty in a photometrically consistent and differentiable form~(\cite{yu2025hammer, yugay2025magic}). This representation supports submap-based pipelines, where each agent maintains locally optimized submaps while still allowing global fusion through consensus or hierarchical updates~(\cite{thomas2025grand}). Moreover, differentiable rendering loops tightly couple geometry and photometric feedback, enabling gradient-based map refinement across agents. Collectively, these motifs define the core structural pattern of scalable and photorealistic collaborative neural mapping.

A second cross-cutting pattern lies in the integration of semantic distillation and knowledge preservation within the Gaussian representation. Systems such as HAMMER and MAGiC-SLAM embed semantic features from pretrained vision–language encoders (\textit{e.g.}, CLIP, DINOv2) directly into the Gaussian parameters. This allows the shared map to function not only as a geometric model but also as a semantic knowledge field, supporting tasks like language-guided navigation and object-level reasoning. In these architectures, however, semantic distillation is managed through centralized supervision: a server aggregates streamed Gaussian updates from multiple agents, performs continuous optimization of the global 3DGS model, and redistributes refined parameters. This partially centralized pipeline offloads computation and enforces coherence across agents, ensuring a unified global latent space. In contrast, fully distributed implicit systems such as MNE-SLAM~(\cite{deng2025mne}) and MACIM, achieve online knowledge transfer through peer-to-peer distillation and consensus optimization, which use a teacher-student model where one agent's network learns to reproduce the renderings of another's by minimizing a distillation loss over rendered RGB and depth outputs. These methods preserve global coherence under bandwidth constraints, demonstrating that shared Gaussian or implicit representations can function as vehicles for distributed learning, even without centralized supervision.

\subsubsection{Inter-agent correspondence and consistency mechanisms:}
To address the critical challenges of global consistency and radiance alignment, CoGS-SLAM systems must ensure that independently optimized Gaussian maps are not only geometrically aligned but also photometrically coherent when fused. Current approaches often employ multi-stage mechanisms combining classical geometric methods with novel radiance-based techniques. After coarse geometric alignment, specific strategies are needed to merge the Gaussian primitives themselves while ensuring visual coherence. Several patterns have emerged:
\begin{itemize}
    \item Semantic-Guided Registration: Methods like SIREN (\cite{shorinwa2025siren}) perform registration without requiring an initial pose estimate by leveraging semantic features embedded in the Gaussians to find robust correspondences, followed by geometric and photometric refinement. 
    \item Gaussian Consensus Protocols: MAC-Ego3D~(\cite{xu2025mac} jointly optimize the poses of all agents and the parameter of a single, shared global Gaussian map by minimizing a collaborative photometric loss, ensuring inherent consistency.
    \item Deformable Models and Refinement: MAGiC-SLAM uses a deformable representation for Gaussians, allowing submaps to be non-rigidly warped after PGO to correct residual misalignments and eliminate visual seams during fusion. 
\end{itemize}
In addition, distributed neural SLAMs often rely on peer-to-peer knowledge distillation (\textit{e.g.}, MNE-SLAM, MCN-SLAM) or inter-loop closure losses to align submaps asynchronously, minimizing direct communication of dense map data. These approaches highlight that consistency in multi-agent settings is a continuous negotiation of geometry, appearance, and trajectories across agents.

Interestingly, many of these consistency challenges mirror those in single-agent GS-SLAM, where correcting pose drift, handling unstable splats, and reconciling photometric variations are also crucial. Systems such as NEDS-SLAM~(\cite{ji2024neds}) use virtual view pruning to remove inconsistent Gaussians, while Splat-SLAM and NEWTON~(\cite{matsuki2024newton}) introduce deformable or view-centric submaps to absorb global pose corrections. These mechanisms suggest promising directions for multi-agent systems: \textit{e.g.}, view-consistency pruning could be extended to filter unreliable inter-agent correspondences, deformable Gaussian maps could support online alignment under drift, and photometric regularization could reduce appearance conflicts across heterogeneous robots. Thus, solutions first explored for single-agent GS-SLAM already provide a blueprint for tackling the harder problem of multi-agent global consistency.

Overall, inter-agent consistency in CoGS-SLAM significantly complicates classical Co-SLAM due to the dense, photorealistic nature of Gaussian representations. Emerging systems increasingly converge toward hybrid designs where consensus optimization, semantic registration, and photometric regularization collectively ensure globally consistent multi-robot 3DGS mapping.

\subsubsection{Asynchronous fusion and pose optimization:}
A key design pattern enabling scalability and robustness in CoGS-SLAM in asynchronous operation. This allows individual agents to perform local map updates, communicate, and participate in global optimization without requiring strict temporal synchronization across the entire team. Asynchrony is crucial for mitigating latency and bandwidth bottlenecks inherent in multi-robot systems operating under variable communication conditions. In practice, asynchronous mechanisms decouple local map updates from global alignment processes. Frameworks like MCN-SLAM~(\cite{deng2025mcn}) and MNE-SLAM~(\cite{deng2025mne}) exemplify this by allowing agents to update their local implicit features independently. Inter-agent fusion occurs opportunistically when spatial overlap is detected through rendering-based consistency checks performed only when communication is available. Similarly, explicit 3DGS frameworks often rely on asynchronous updates. MAGiC-SLAM~(\cite{yugay2025magic}) maintains each agent's submaps that are fused only when inter-agent loop closures are detected and verified. MAC-Ego3D uses an asynchronous multi-agent consensus mechanism, allowing agents to contribute updates to a shared global map representation without waiting for others. In centralized systems like HAMMER, agents stream data asynchronously to the server, which continuously refines the global map and poses without halting agent operations.

Asynchrony, therefore, acts as a critical system-level pattern. It does not fundamentally change the optimization objective (\textit{e.g.}, minimizing geometric or photometric error) but rather dictates how and when agents contribute to and benefit from the global optimization process. This temporal flexibility is essential for achieving distributed scalability and maintaining real-time responsiveness in dynamic multi-robot scenarios, complementing the spatial consistency mechanisms discussed previously.


\subsubsection{Compression-oriented efficiency for real-time scalability:}
Addressing communication and memory bottlenecks is paramount for the real-time scalability of CoGS-SLAM systems. This involves adopting data-efficient map representations and optimizing communication protocols to minimize data transfer during collaboration. Instead of traditional memory-intensive structures like dense voxel grids, CoGS-SLAM systems leverage compact representations to reduce storage and transmission costs. Systems using implicit neural fields, such as MCN-SLAM~(\cite{deng2025mcn}) and MNE-SLAM~(\cite{deng2025mne}), employ hybrid encodings (\textit{e.g.}, planar-grid-coordinate or multi-resolution hash grids) that often scale quadratically or better with scene size, rather than cubically like standard grids. CP-SLAM~(\cite{liu2025cpl}) uses a neural point-based representation where each explicit 3D point carries a learnable feature vector. This allows dense mapping with fewer primitives compared to voxels, and grid-based filtering further reduces redundancy. The 3DGS representation itself offers a degree of compactness compared to dense volumetric methods. Furthermore, systems like MAGiC-SLAM~(\cite{yugay2025magic}) and GRAND-SLAM~(\cite{thomas2025grand}) utilize the explicit nature of 3DGS for efficient sub-map storage and can selectively transmit or prune low-opacity or redundant splats, minimizing both VRAM and disk usage. These representation strategies collectively aim to balance computational efficiency with the goal of high-fidelity scene reconstruction.

Complementing these representations, CoGS-SLAM frameworks leverage distributed and peer-to-peer communication collaboration designed to minimize bandwidth usage. Instead of transmitting raw sensor data or full geometric maps, implicit systems like MCN-SLAM and MNE-SLAM exchange only neural network parameters ($\theta$) or keyframe descriptors during optimization or knowledge distillation, achieving significant communication efficiency gains over centralized designs. DCL-SLAM~(\cite{zhong_dcl-slam_2023}) implements a three-stage distributed loop-closure protocol progressing from lightweight global descriptors to filtered point clouds and finally to verified relative poses to further limit data transmission to essential information. Neural implicit approaches such as Di-NeRF extend this principle by sharing compact NeRF weights over mesh networks, avoiding the cost of raw visual data exchange. These communication strategies collectively ensure scalable and bandwidth-efficient collaboration between agents while maintaining global map consistency.

\section{Comparative Analysis and Applications}
\label{sec:analysis}
This section presents a comparative analysis of existing CoGS-SLAM frameworks, evaluating their performance using standard benchmarks and datasets across various metrics. It further outlines practical applications where these systems demonstrate robustness, scalability, and real-world adaptability.
\subsection{Benchmarks and Datasets}

Photorealistic synthetic datasets provide controlled environments with perfect ground truth for geometry and camera poses, making them ideal for initial validation and ablation studies. The Replica dataset and its multi-agent extension, Replica-MultiAgent~(\cite{straub2019replica}), are the most widely used synthetic benchmarks. Replica-MultiAgent provided synchronized RGB-D trajectories for two collaborating agents in indoor scenes (Office-0, Apartment-0/1/2), each containing 1.5–2.5k frames. It has served as a primary evaluation tool for numerous recent methods, including HAMMER, CP-SLAM, MAGiC-SLAM, and GRAND-SLAM. A related dataset, ReplicaCAD~(\cite{szot2021habitat}), derived from the same engine, is adopted by MACIM for evaluation in Apartment-2 and Apartment-3.

On the other hand, real-world benchmarks are crucial for evaluating robustness against sensor noise, lighting variations, complex scene dynamics, and less accurate ground truth. ScanNet~(\cite{dai2017scannet}) provides large-scale indoor RGB-D sequences ($>5,000$ frames) and is used by MACIM for six distinct scenes. TUM RGB-D~(\cite{sturm2012benchmark}) remains a classic choice for camera tracking and localization evaluation, as employed in CP-SLAM and MCN-SLAM~(\cite{liu2025cpl, deng2025mcn}). Similarly, the 7-Scenes dataset~(\cite{glocker2013real}) supports egocentric RGB-D tracking for MAC-Ego3D, while GRAND-SLAM also tests scalability on the Kimera-Multi Outdoor dataset~(\cite{chang2021kimera}), spanning over $1.8$ km of multi-agent trajectories. In addition, several recent studies have introduced novel datasets tailored for evaluating collaborative and neural SLAM systems. LAMP 2.0~(\cite{chang2022lamp}) benchmarks multi-robot mapping across diverse DARPA Subterranean Challenge environments—Tunnel, Urban, Finals, and Kentucky Underground—featuring heterogeneous robot teams operating under extreme conditions. MNE-SLAM presents the Indoor Neural SLAM (INS) dataset, offering high-fidelity 3D mesh ground truth and continuous-time camera trajectories across indoor environments exceeding $1,000$ m$^{2}$. Complementarily, MCN-SLAM contributes the Dense SLAM (DES) dataset, the first large-scale neural dense SLAM benchmark spanning both compact indoor scenes and extensive ($>276,000$ m$^{2}$) outdoor areas.

Beyond neural or Gaussian-splatting datasets, several multi-robot SLAM benchmarks provide valuable context for evaluating coordination, perception, and robustness in heterogeneous teams. UTIAS~(\cite{leung2011utias}) represents one of the earliest cooperative localization datasets, featuring five monocular robots in an indoor workspace with range–bearing sensing. AirMuseum~(\cite{dubois2020airmuseum}) focuses on stereo-visual–inertial C-SLAM, including three ground robots and a drone with explicit inter-robot detection via AprilTags. S3E~(\cite{feng2024s3e}) contributes multimodal data (360$^{\circ}$ LiDAR, stereo, IMU, UWB) across controlled collaborative trajectories, enabling study of overlap sensitivity. SubT-MRS~(\cite{zhao2024subt}) and DiTer++~(\cite{kim2025diter++}) extend this to challenging environments, integrating aerial, legged, and wheeled robots with multimodal perception (LiDAR, RGB, thermal). Additional resources such as FordAV, GRACO~(\cite{zhu2023graco}), and SubT derivatives like NeBula and CERBERUS further broaden C-SLAM evaluation.

Despite this growing ecosystem, a significant gap remains: none of the existing datasets fully capture the unique requirements for benchmarking CoGS-SLAM, particularly concerning photorealistic fidelity, semantic consistency, or bandwidth-aware constraints. Developing standardized multi-agent datasets tailored to these specific challenges remains a key open task.

\subsection{Metrics}
The evaluation of CoGS-SLAM systems typically spans four complementary dimensions: localization accuracy, photorealistic quality, system performance, and semantic consistency.

\textit{(1) Localization accuracy}: Standard SLAM metrics are used, primarily the Absolute Trajectory Error (ATE-RMSE) or the Relative Position Error (RPE). These are reported by most recent works, including GRAND-SLAM, MaAGiC-SLAM, CP-SLAM, MCN-SLAM, and MNE-SLAM, in which GRAND-SLAM achieves lower ATE through robust pose graph optimization and MAGiC-SLAM emphasizes inter-agent loop closure.
    
\textit{(2) Photorealistic quality}: The quality of novel view synthesis is assessed by comparing rendered views to ground-truth images using standard image quality metric: PSNR, SSIM~(\cite{wang2004image}), or perceptual metrics such as LPIPS~(\cite{zhang2018unreasonable}). GS-based systems generally outperform mesh or voxel baselines, but alignment errors can substantially degrade photometric fidelity.
    
\textit{(3) System performance}: Real-time feasibility and resource usage are critical. Key metrics include FPS, memory footprint, communication bandwidth, and latency. MAC-Ego3D demonstrates reduced bandwidth via consensus updates, and HAMMER quantifies semantic query latency.
    
\textit{(4) Semantic consistency (emerging)}: As semantic mapping becomes integrated, metrics are needed to evaluate the consistency and accuracy of semantic labels. HAMMER uniquely introduces semantic retrieval accuracy (\textit{e.g.}, mean recall for open-vocabulary queries) as a metric, representing a new dimension beyond geometric and appearance.

Beyond quantitative metrics, which are detailed in Table~\ref{tab:comp}, visual inspection remains vital for judging perceptual and structural integrity. Examining rendered scenes, trajectory overlays, and mesh/splat visualizations helps identify subtle issues like floaters, discontinuities, seams, or misalignments that may not be fully captured by average scores. While these metrics provide complementary perspectives, they are rarely reported jointly in the current literature. Most works focus primarily on trajectory accuracy or reconstruction quality. However, recent distributed systems like MNE-SLAM, MCN-SLAM have begun to analyze localization, rendering, and communication efficiency together, setting a trend toward more holistic evaluation.
\begin{table*}[!ht]
    \scriptsize
    \centering
    \caption{Qualitative comparison on the Multiagent Replica dataset, best and second best values are highlighted in yellow and green, respectively.}
    \begin{tabular}{>{\centering\arraybackslash}p{2.8cm}|>{\centering\arraybackslash}p{2.5cm}>{\centering\arraybackslash}p{2.5cm}>{\centering\arraybackslash}p{2.5cm}>{\centering\arraybackslash}p{2.5cm}>{\arraybackslash}p{2.5cm}}
    \toprule
    \multirow{2}{*}{Methods} & \centering Office 0 & \centering Apartment-0 & \centering Apartment-1 & \centering Apartment-2 & Average \\ 
    & \multicolumn{5}{c}{RMSE[cm]$\downarrow$/PSNR[dB]$\uparrow$/SSIM$\uparrow$/LPIPS$\downarrow$/Depth L1[cm]$\downarrow$/Time[min]$\downarrow$} \\  \midrule
    CP-SLAM (\cite{hu2023cp})    & 0.65/28.56/0.87 /0.29/2.74/200+ & 0.95/26.12/0.79 /0.41/19.93/200+ & 1.42/12.16/0.31 /0.97/66.77/200+ & 1.91/23.98/0.81 /0.39/2.47/200+ & 1.23/22.71/0.69 /0.51/22.98/200+ \\
    MAGiC-SLAM (\cite{yugay2025magic})& \hlgr{0.27}/39.32/\hl{0.99} /0.05/\hlgr{0.41}/\hlgr{85}   & \hlgr{0.16}/35.96/\hlgr{0.98} /0.09/0.64/146   & \hlgr{0.26}/30.01/0.95 /0.18/3.16/\hlgr{158}   & 0.32/30.01/0.95 /0.18/3.16/\hlgr{144}  & 0.25/34.26/0.97 /0.13/1.30/\hlgr{133} \\
    MAC-Ego3D (\cite{xu2025mac}) & \hl{0.14}/42.43/\hlgr{0.98} /\hl{0.03}/\hlgr{0.41}/---  & \hl{0.13}/\hlgr{42.83}/\hlgr{0.98} /\hlgr{0.05}/\hlgr{0.48}/---   & \hl{0.19}/\hlgr{36.28}/\hlgr{0.97} /\hlgr{0.06}/\hlgr{1.03}/---   & \hl{0.10}/\hlgr{38.59}/\hlgr{0.98} /\hlgr{0.06}/\hlgr{1.03}/---  & \hl{0.14}/\hlgr{40.04}/\hlgr{0.98} /\hlgr{0.05}/\hlgr{0.74}/--- \\
    GRAND-SLAM (\cite{thomas2025grand})& \hlgr{0.27}/\hl{43.12}/\hl{0.99} /\hl{0.03}/\hl{0.25}/---  & 0.23/\hl{44.15}/\hl{0.99} /\hl{0.03}/\hl{0.31}/---   & 0.32/\hl{38.65}/\hl{0.99} /\hl{0.05}/\hl{0.77}/---   & \hlgr{0.18}/\hl{39.46}/\hl{0.99} /\hl{0.05}/\hl{0.29}/---  & \hlgr{0.25}/\hl{41.35}/\hl{0.99} /\hl{0.04}/\hl{0.41}/--- \\
    HAMMER  (\cite{yu2025hammer})   & ---/\hlgr{43.01}/\hlgr{0.98}  /\hlgr{0.04}/0.94/\hl{8}    &  ---/40.89/0.97 /0.06/1.27/\hl{8}     &  ---/31.43/0.90 /0.13/2.40/\hl{8}     &  ---/33.78/0.94 /0.11/1.25/\hl{8}    &  ---/37.28/0.96 /0.09/1.47/\hl{8}   \\
    \bottomrule
    \end{tabular}
    \label{tab:comp}
\end{table*}
\subsection{Applications}
The ability of CoGS-SLAM systems to generate high-fidelity, large-scale maps opens up numerous application possibilities for multi-robot teams.
\begin{itemize}
    \item Disaster response: Multi-robot CoGS-SLAM systems can rapidly survey hazardous environments (\textit{e.g.}, collapsed buildings, mines, or industrial accidents) to generate dense and photorealistic reconstructions. The fine-grained textures of Gaussian maps allow remote experts to identify material stress, cracks, and obstructions that LiDAR-only reconstructions might miss, enhancing both situational awareness and mission planning~(\cite{lei2024gart, chen2025splat}).
    
    \item Large-Scale Digital Twinning and Surveying: CoGS-SLAM enables the rapid creation of detailed 3D models of large environments, such as construction sites, factories, or urban areas. These ``digital twins'' can be used for monitoring, planning, and simulation~(\cite{yu2025hammer}).
    
    \item Collaborative exploration and Active Reconstruction: Teams of robots can autonomously explore and map unknown environments more efficiently than a single robot. Active reconstruction strategies, guided by metrics like semantic uncertainty, can direct robots to gather the most informative views, leading to higher-quality maps with less effort.
    
    \item Heterogeneous Team Operations: Centralized architectures like HAMMER are specifically designed to integrate data from diverse robot platforms (\textit{e.g.}, ground robots, drones, AR glasses), enabling complex collaborative tasks where different robots contribute unique perspectives or sensor data.    
    \item Shared Augmented/Virtual Reality (AR/VR): High-fidelity GS maps provide a shared spatial context for immersive AR/VR experiences, allowing multiple users or robots to interact within a consistent virtual representation of the real world.
    
    \item Semantic Interaction: Maps augmented with semantic information, as produced by HAMMER, allow for higher-level interactions, such as instructing robots using natural language queries (\textit{e.g.}, ``Inspect the red container'').
\end{itemize}

\section{Open Questions and Research Directions}
\label{sec:open}
Despite the rapid progress in this survey, CoGS-SLAM remains an emerging field with fewer than a dozen representative systems. This section highlights feasibility, but leaves many fundamental questions open, and outlines several directions for future research. 

\textit{(1) Real-time synchronization and consistency:} Current systems often struggle with latency, asynchronous data arrival, and maintaining photometric consistency across submaps fused from different agents or time instances, especially under limited bandwidth. Geometric alignment alone does not guarantee seamless visual appearance. Future work should explore predictive synchronization techniques and hierarchical fusion schemes that explicitly model and compensate for temporal misalignments. Research into uncertainty-aware radiance fusion, potentially weighting contributions based on estimated pose uncertainty or temporal proximity, could improve visual coherence. Developing adaptive communication scheduling protocols that prioritize updates based on their expected impact on global consistency is also crucial.

\textit{(2) Robust inter-agent loop closure and data association:} Reliably detecting loop closures between agents, especially across different viewpoints, sensor modalities (heterogeneity), or significant appearance changes (\textit{e.g.}, lighting, weather), remains difficult. Current reliance on classical descriptors (ORB, NetVLAD) or even standard learned descriptors can be brittle. Leveraging foundation model-based descriptors (\textit{e.g.}, from CLIP, DINO) offers a promising avenue due to their inherent robustness to appearance variation. Combining these semantic descriptors with photometric validation using the differentiable renderer could create highly robust multi-stage loop closure verification pipelines. Furthermore, developing robust outlier rejection techniques specifically tailored for the high dimensionality and potential ambiguity of GS features is needed.

\textit{(3) Semantic-aware and instance-level mapping:}  
While systems like HAMMER demonstrate the potential of semantic GS maps for tasks like open-vocabulary queries, ensuring semantic reliability, instance consistency (distinguishing between multiple objects of the same class), and accurate segmentation boundaries remains challenging. Research should focus on enhancing semantic fidelity by integrating online open-vocabulary segmentation (potentially using models like SAM) with instance uncertainty estimation. This could guide active reconstruction strategies. Explicit instance-level grouping of Gaussians and techniques for semantic distillation across agents could refine low-quality object regions and reduce label ambiguity. Exploring lightweight foundation model integration for real-time semantic mapping is also critical.

\textit{(4) Multi-modal fusion for Robustnesss:}  
Most CoGS-SLAM systems are camera-centric, limiting their robustness in challenging conditions like poor lighting, textureless environments, or adverse weather where cameras struggle. Extending CoGS-SLAM to incorporate multi-modal data is essential. This includes LiDAR-seeded Gaussian initialization to improve geometry, inertial-assisted tracking and optimization for better motion estimation, and potentially integrating thermal or radar data for all-weather operation. Developing fusion strategies that effectively combine the strengths of different sensor modalities within the GS framework is a key area for future work.

\textit{(5) Lifelong learning and continual adaptation:}
Current CoGS-SLAM systems typically focus on building maps during a single deployment. However, real-world applications require persistent digital twins that can be continuously updated over time, adapt to environmental changes (\textit{e.g.}, moved furniture, seasonal variations), and potentially forget outdated information. Research is needed on efficient online map updates for 3DGS, including effective change detection mechanisms. Developing forgetting strategies to manage map size and remove outdated information, possibly guided by semantics or temporal cues, is crucial for lifelong operation. Exploring transfer learning techniques to adapt maps between different times or related environments also presents an open challenge.

\textit{(6) Scalability and large-scale deployments:}
Scaling CoGS-SLAM beyond a few robots in relatively constrained environments remains difficult. Centralized systems face server bottlenecks, while distributed systems require robust protocols for maintaining consistency under intermittent communication. Further research into optimal submap partitioning strategies is needed. Developing minimal communication statistics that agents can exchange to maintain global consistency without transmitting full maps is critical for distributed approaches. Designing consensus algorithms specifically for distributed GS optimization that are robust to intermittent network connectivity and packet loss is essential for large-scale deployments in urban or disaster scenarios.

\textit{(7) Sim2Real generalization and robustness:}
Many CoGS-SLAM systems are primarily evaluated on synthetic datasets, which lack the sensor noise, lighting variability, motion blur, and calibration errors present in real-world deployments. Focused efforts are required on domain adaptation techniques for CoGS-SLAM. Incorporating uncertainty-aware parameterization for Gaussians could help model sensor noise and improve robustness. Rigorous evaluation on diverse, large-scale real-world datasets under challenging conditions is necessary to truly validate system performance and drive progress toward practical applications.

\section{Conclusion} \label{sec:conclusion}
This survey has provided a comprehensive review of the rapidly evolving field of CoGS-SLAM. We presented a taxonomy categorizing current systems based on their architecture (centralized vs. distributed) and analyzed the core technical sub-problems: multi-agent consistency and alignment, and communication-efficient collaboration optimization. Our review identified recurring design patterns, including the shared Gaussian representations, inter-agent correspondence, asynchronous fusion and pose optimization, and real-time scalability. Despite significant progress, persistent challenges hinder widespread adoption. Ensuring multi-agent consistency—both geometrically and photometrically—demands robust Gaussian submap fusion techniques that avoid visual artifacts. The high sensitivity of 3DGS rendering to pose accuracy makes robust loop closure crucial, yet current methods often rely on descriptors not specifically tailored for Gaussian representations. Furthermore, maintaining real-time performance and scalability across large deployments requires further advancements in optimization and communication efficiency.

Future research should prioritize several key areas: (i) developing real-time synchronization methods and communication efficient data sharing protocols; (ii) creating more robust inter-agent loop closure techniques, potentially using foundation models and photometric validation specifically designed for 3DGS; (iii) enhancing robustness through multi-modal fusion; and (iv) advancing toward scalable, lifelong mapping systems capable.

In conclusion, CoGS-SLAM represents a promising paradigm shift in multi-robot perception. While the field is rapidly maturing, realizing its full potential in robotics, disaster response, AR/VR, and large-scale digital twinning hinges on effectively addressing the fundamental challenges of consistency, communication efficiency, and scalability.


\section*{DECLARATION OF GENERATIVE AI AND AI-ASSISTED TECHNOLOGIES IN THE WRITING PROCESS}
During the preparation of this work the author(s) used Gemini in order to check grammar typos and improve scientific writing. After using this tool/service, the author(s) reviewed and edited the content as needed and take(s) full responsibility for the content of the publication.

\bibliography{ifacconf}             

\begin{thebibliography}{68}
\providecommand{\natexlab}[1]{#1}
\providecommand{\url}[1]{\texttt{#1}}
\providecommand{\urlprefix}{URL }
\expandafter\ifx\csname urlstyle\endcsname\relax
  \providecommand{\doi}[1]{doi:\discretionary{}{}{}#1}\else
  \providecommand{\doi}{doi:\discretionary{}{}{}\begingroup \urlstyle{rm}\Url}\fi

\bibitem[{Asadi et~al.(2024)Asadi, Zareinia, and Saeedi}]{asadi2024di}
Asadi, M., Zareinia, K., and Saeedi, S. (2024).
\newblock Di-nerf: Distributed nerf for collaborative learning with relative pose refinement.
\newblock \emph{IEEE Robotics and Automation Letters}.

\bibitem[{Bao et~al.(2025)Bao, Ding, Huo, Liu, Li, Li, Gao, and Luo}]{bao20253d}
Bao, Y., Ding, T., Huo, J., Liu, Y., Li, Y., Li, W., Gao, Y., and Luo, J. (2025).
\newblock 3d gaussian splatting: Survey, technologies, challenges, and opportunities.
\newblock \emph{IEEE Transactions on Circuits and Systems for Video Technology}.

\bibitem[{Canh et~al.(2023)Canh, Elibol, Chong, and HoangVan}]{canh2023object}
Canh, T.N., Elibol, A., Chong, N.Y., and HoangVan, X. (2023).
\newblock Object-oriented semantic mapping for reliable uavs navigation.
\newblock In \emph{2023 12th International Conference on Control, Automation and Information Sciences (ICCAIS)}, 139--144. IEEE.

\bibitem[{Canh et~al.(2024)Canh, Nguyen, HoangVan, Elibol, and Chong}]{canh2024s3m}
Canh, T.N., Nguyen, V.T., HoangVan, X., Elibol, A., and Chong, N.Y. (2024).
\newblock S3m: Semantic segmentation sparse mapping for uavs with rgb-d camera.
\newblock In \emph{2024 IEEE/SICE International Symposium on System Integration (SII)}, 899--905. IEEE.

\bibitem[{Cao and Beltrame(2021)}]{cao2021vir}
Cao, Y. and Beltrame, G. (2021).
\newblock Vir-slam: Visual, inertial, and ranging slam for single and multi-robot systems.
\newblock \emph{Autonomous Robots}, 45(6), 905--917.

\bibitem[{Castle et~al.(2008)Castle, Klein, and Murray}]{castle2008video}
Castle, R., Klein, G., and Murray, D.W. (2008).
\newblock Video-rate localization in multiple maps for wearable augmented reality.
\newblock In \emph{2008 12th IEEE International Symposium on Wearable Computers}, 15--22. IEEE.

\bibitem[{Chang et~al.(2022)Chang, Ebadi, Denniston, Ginting, Rosinol, Reinke, Palieri, Shi, Chatterjee, Morrell et~al.}]{chang2022lamp}
Chang, Y., Ebadi, K., Denniston, C.E., Ginting, M.F., Rosinol, A., Reinke, A., Palieri, M., Shi, J., Chatterjee, A., Morrell, B., et~al. (2022).
\newblock Lamp 2.0: A robust multi-robot slam system for operation in challenging large-scale underground environments.
\newblock \emph{IEEE Robotics and Automation Letters}, 7(4), 9175--9182.

\bibitem[{Chang et~al.(2021)Chang, Tian, How, and Carlone}]{chang2021kimera}
Chang, Y., Tian, Y., How, J.P., and Carlone, L. (2021).
\newblock Kimera-multi: a system for distributed multi-robot metric-semantic simultaneous localization and mapping.
\newblock In \emph{2021 IEEE International Conference on Robotics and Automation (ICRA)}, 11210--11218. IEEE.

\bibitem[{Charatan et~al.(2024)Charatan, Li, Tagliasacchi, and Sitzmann}]{charatan2024pixelsplat}
Charatan, D., Li, S.L., Tagliasacchi, A., and Sitzmann, V. (2024).
\newblock pixelsplat: 3d gaussian splats from image pairs for scalable generalizable 3d reconstruction.
\newblock In \emph{Proceedings of the IEEE/CVF conference on computer vision and pattern recognition}, 19457--19467.

\bibitem[{Chen et~al.(2025)Chen, Shorinwa, Bruno, Swann, Yu, Zeng, Nagami, Dames, and Schwager}]{chen2025splat}
Chen, T., Shorinwa, O., Bruno, J., Swann, A., Yu, J., Zeng, W., Nagami, K., Dames, P., and Schwager, M. (2025).
\newblock Splat-nav: Safe real-time robot navigation in gaussian splatting maps.
\newblock \emph{IEEE Transactions on Robotics}.

\bibitem[{Dai et~al.(2017)Dai, Chang, Savva, Halber, Funkhouser, and Nie{\ss}ner}]{dai2017scannet}
Dai, A., Chang, A.X., Savva, M., Halber, M., Funkhouser, T., and Nie{\ss}ner, M. (2017).
\newblock Scannet: Richly-annotated 3d reconstructions of indoor scenes.
\newblock In \emph{Proceedings of the IEEE conference on computer vision and pattern recognition}, 5828--5839.

\bibitem[{Deng et~al.(2025{\natexlab{a}})Deng, Shen, Chen, Yuan, Shen, Peng, Wu, Wang, Xie, Wang et~al.}]{deng2025mcn}
Deng, T., Shen, G., Chen, X., Yuan, S., Shen, H., Peng, G., Wu, Z., Wang, J., Xie, L., Wang, D., et~al. (2025{\natexlab{a}}).
\newblock Mcn-slam: Multi-agent collaborative neural slam with hybrid implicit neural scene representation.
\newblock \emph{arXiv preprint arXiv:2506.18678}.

\bibitem[{Deng et~al.(2025{\natexlab{b}})Deng, Shen, Xun, Yuan, Jin, Shen, Wang, Wang, Wang, Wang et~al.}]{deng2025mne}
Deng, T., Shen, G., Xun, C., Yuan, S., Jin, T., Shen, H., Wang, Y., Wang, J., Wang, H., Wang, D., et~al. (2025{\natexlab{b}}).
\newblock Mne-slam: Multi-agent neural slam for mobile robots.
\newblock In \emph{Proceedings of the Computer Vision and Pattern Recognition Conference}, 1485--1494.

\bibitem[{Deng et~al.(2024)Deng, Tang, Yang, Wang, and Yue}]{deng2024macim}
Deng, Y., Tang, Y., Yang, Y., Wang, D., and Yue, Y. (2024).
\newblock Macim: Multi-agent collaborative implicit mapping.
\newblock \emph{IEEE Robotics and Automation Letters}, 9(5), 4369--4376.

\bibitem[{Dissanayake et~al.(2001)Dissanayake, Newman, Clark, Durrant-Whyte, and Csorba}]{dissanayake2001solution}
Dissanayake, M.G., Newman, P., Clark, S., Durrant-Whyte, H.F., and Csorba, M. (2001).
\newblock A solution to the simultaneous localization and map building (slam) problem.
\newblock \emph{IEEE Transactions on robotics and automation}, 17(3), 229--241.

\bibitem[{Dubois et~al.(2020)Dubois, Eudes, and Fr{\'e}mont}]{dubois2020airmuseum}
Dubois, R., Eudes, A., and Fr{\'e}mont, V. (2020).
\newblock Airmuseum: a heterogeneous multi-robot dataset for stereo-visual and inertial simultaneous localization and mapping.
\newblock In \emph{2020 IEEE International Conference on Multisensor Fusion and Integration for Intelligent Systems (MFI)}, 166--172. IEEE.

\bibitem[{Feng et~al.(2024)Feng, Qi, Zhong, Chen, Chen, Chen, Wu, and Ma}]{feng2024s3e}
Feng, D., Qi, Y., Zhong, S., Chen, Z., Chen, Q., Chen, H., Wu, J., and Ma, J. (2024).
\newblock S3e: A multi-robot multimodal dataset for collaborative slam.
\newblock \emph{IEEE Robotics and Automation Letters}.

\bibitem[{Forster et~al.(2013)Forster, Lynen, Kneip, and Scaramuzza}]{forster2013collaborative}
Forster, C., Lynen, S., Kneip, L., and Scaramuzza, D. (2013).
\newblock Collaborative monocular slam with multiple micro aerial vehicles.
\newblock In \emph{2013 IEEE/RSJ International Conference on Intelligent Robots and Systems}, 3962--3970. IEEE.

\bibitem[{Glocker et~al.(2013)Glocker, Izadi, Shotton, and Criminisi}]{glocker2013real}
Glocker, B., Izadi, S., Shotton, J., and Criminisi, A. (2013).
\newblock Real-time rgb-d camera relocalization.
\newblock In \emph{2013 IEEE International Symposium on Mixed and Augmented Reality (ISMAR)}, 173--179. IEEE.

\bibitem[{Hu et~al.(2023)Hu, Mao, Bao, Zhang, and Cui}]{hu2023cp}
Hu, J., Mao, M., Bao, H., Zhang, G., and Cui, Z. (2023).
\newblock Cp-slam: Collaborative neural point-based slam system.
\newblock \emph{Advances in Neural Information Processing Systems}, 36, 39429--39442.

\bibitem[{Huang et~al.(2022)Huang, Shan, Chen, and Englot}]{huang2021disco}
Huang, Y., Shan, T., Chen, F., and Englot, B. (2022).
\newblock Disco-slam: Distributed scan context-enabled multi-robot lidar slam with two-stage global-local graph optimization.
\newblock \emph{IEEE Robotics and Automation Letters}, 7(2), 1150--1157.

\bibitem[{Jamieson et~al.(2021)Jamieson, Fathian, Khosoussi, How, and Girdhar}]{jamieson2021multi}
Jamieson, S., Fathian, K., Khosoussi, K., How, J.P., and Girdhar, Y. (2021).
\newblock Multi-robot distributed semantic mapping in unfamiliar environments through online matching of learned representations.
\newblock In \emph{2021 IEEE international conference on robotics and automation (ICRA)}, 8587--8593. IEEE.

\bibitem[{Ji et~al.(2024)Ji, Liu, Xie, Ma, Xie, and Liu}]{ji2024neds}
Ji, Y., Liu, Y., Xie, G., Ma, B., Xie, Z., and Liu, H. (2024).
\newblock Neds-slam: A neural explicit dense semantic slam framework using 3d gaussian splatting.
\newblock \emph{IEEE Robotics and Automation Letters}.

\bibitem[{Karrer et~al.(2018)Karrer, Schmuck, and Chli}]{karrer2018cvi}
Karrer, M., Schmuck, P., and Chli, M. (2018).
\newblock Cvi-slam—collaborative visual-inertial slam.
\newblock \emph{IEEE Robotics and Automation Letters}, 3(4), 2762--2769.

\bibitem[{Kazerouni et~al.(2022)Kazerouni, Fitzgerald, Dooly, and Toal}]{kazerouni2022survey}
Kazerouni, I.A., Fitzgerald, L., Dooly, G., and Toal, D. (2022).
\newblock A survey of state-of-the-art on visual slam.
\newblock \emph{Expert Systems with Applications}, 205, 117734.

\bibitem[{Keetha et~al.(2024)Keetha, Karhade, Jatavallabhula, Yang, Scherer, Ramanan, and Luiten}]{keetha2024splatam}
Keetha, N., Karhade, J., Jatavallabhula, K.M., Yang, G., Scherer, S., Ramanan, D., and Luiten, J. (2024).
\newblock Splatam: Splat track \& map 3d gaussians for dense rgb-d slam.
\newblock In \emph{Proceedings of the IEEE/CVF Conference on Computer Vision and Pattern Recognition}, 21357--21366.

\bibitem[{Kim et~al.(2025)Kim, Kim, Jeong, Shin, and Cho}]{kim2025diter++}
Kim, J., Kim, H., Jeong, S., Shin, Y., and Cho, Y. (2025).
\newblock Diter++: Diverse terrain and multi-modal dataset for multi-robot slam in multi-session environments.
\newblock In \emph{2025 IEEE International Conference on Robotics and Automation (ICRA)}, 12187--12193. IEEE.

\bibitem[{K{\"u}mmerle et~al.(2011)K{\"u}mmerle, Grisetti, Strasdat, Konolige, and Burgard}]{kummerle2011g}
K{\"u}mmerle, R., Grisetti, G., Strasdat, H., Konolige, K., and Burgard, W. (2011).
\newblock g 2 o: A general framework for graph optimization.
\newblock In \emph{2011 IEEE international conference on robotics and automation}, 3607--3613. IEEE.

\bibitem[{Lajoie et~al.(2020)Lajoie, Ramtoula, Chang, Carlone, and Beltrame}]{lajoie2020door}
Lajoie, P.Y., Ramtoula, B., Chang, Y., Carlone, L., and Beltrame, G. (2020).
\newblock Door-slam: Distributed, online, and outlier resilient slam for robotic teams.
\newblock \emph{IEEE Robotics and Automation Letters}, 5(2), 1656--1663.

\bibitem[{Lajoie et~al.(2025)Lajoie, Ramtoula, De~Martini, and Beltrame}]{lajoie20253d}
Lajoie, P.Y., Ramtoula, B., De~Martini, D., and Beltrame, G. (2025).
\newblock 3d foundation model-based loop closing for decentralized collaborative slam.
\newblock \emph{IEEE Robotics and Automation Letters}.

\bibitem[{Lei et~al.(2024)Lei, Wang, Pavlakos, Liu, and Daniilidis}]{lei2024gart}
Lei, J., Wang, Y., Pavlakos, G., Liu, L., and Daniilidis, K. (2024).
\newblock Gart: Gaussian articulated template models.
\newblock In \emph{Proceedings of the IEEE/CVF conference on computer vision and pattern recognition}, 19876--19887.

\bibitem[{Leung et~al.(2011)Leung, Halpern, Barfoot, and Liu}]{leung2011utias}
Leung, K.Y., Halpern, Y., Barfoot, T.D., and Liu, H.H. (2011).
\newblock The utias multi-robot cooperative localization and mapping dataset.
\newblock \emph{The International Journal of Robotics Research}, 30(8), 969--974.

\bibitem[{Li et~al.(2024)Li, Guo, Yi, and Hong}]{li2024distributed}
Li, C., Guo, G., Yi, P., and Hong, Y. (2024).
\newblock Distributed pose-graph optimization with multi-level partitioning for multi-robot slam.
\newblock \emph{IEEE Robotics and Automation Letters}, 9(6), 4926--4933.

\bibitem[{Liu et~al.(2025)Liu, Wen, Liu, and Yu}]{liu2025cpl}
Liu, X., Wen, S., Liu, H., and Yu, F.R. (2025).
\newblock Cpl-slam: Centralized collaborative multi-robot visual-inertial slam using point-and-line features.
\newblock \emph{IEEE Internet of Things Journal}.

\bibitem[{Matsuki et~al.(2024)Matsuki, Tateno, Niemeyer, and Tombari}]{matsuki2024newton}
Matsuki, H., Tateno, K., Niemeyer, M., and Tombari, F. (2024).
\newblock Newton: Neural view-centric mapping for on-the-fly large-scale slam.
\newblock \emph{IEEE Robotics and Automation Letters}, 9(4), 3704--3711.

\bibitem[{Mildenhall et~al.(2021)Mildenhall, Srinivasan, Tancik, Barron, Ramamoorthi, and Ng}]{mildenhall2021nerf}
Mildenhall, B., Srinivasan, P.P., Tancik, M., Barron, J.T., Ramamoorthi, R., and Ng, R. (2021).
\newblock Nerf: Representing scenes as neural radiance fields for view synthesis.
\newblock \emph{Communications of the ACM}, 65(1), 99--106.

\bibitem[{Naumann et~al.(2024)Naumann, Xu, Leutenegger, and Zuo}]{naumann2024nerf}
Naumann, J., Xu, B., Leutenegger, S., and Zuo, X. (2024).
\newblock Nerf-vo: Real-time sparse visual odometry with neural radiance fields.
\newblock \emph{IEEE Robotics and Automation Letters}, 9(8), 7278--7285.

\bibitem[{Newcombe et~al.(2011{\natexlab{a}})Newcombe, Izadi, Hilliges, Molyneaux, Kim, Davison, Kohi, Shotton, Hodges, and Fitzgibbon}]{newcombe2011kinectfusion}
Newcombe, R.A., Izadi, S., Hilliges, O., Molyneaux, D., Kim, D., Davison, A.J., Kohi, P., Shotton, J., Hodges, S., and Fitzgibbon, A. (2011{\natexlab{a}}).
\newblock Kinectfusion: Real-time dense surface mapping and tracking.
\newblock In \emph{2011 10th IEEE international symposium on mixed and augmented reality}, 127--136. Ieee.

\bibitem[{Newcombe et~al.(2011{\natexlab{b}})Newcombe, Lovegrove, and Davison}]{newcombe2011dtam}
Newcombe, R.A., Lovegrove, S.J., and Davison, A.J. (2011{\natexlab{b}}).
\newblock Dtam: Dense tracking and mapping in real-time.
\newblock In \emph{2011 international conference on computer vision}, 2320--2327. IEEE.

\bibitem[{Riazuelo et~al.(2014)Riazuelo, Civera, and Montiel}]{riazuelo2014c2tam}
Riazuelo, L., Civera, J., and Montiel, J.M. (2014).
\newblock C2tam: A cloud framework for cooperative tracking and mapping.
\newblock \emph{Robotics and Autonomous Systems}, 62(4), 401--413.

\bibitem[{Sandstr{\"o}m et~al.(2023)Sandstr{\"o}m, Li, Van~Gool, and Oswald}]{sandstrom2023point}
Sandstr{\"o}m, E., Li, Y., Van~Gool, L., and Oswald, M.R. (2023).
\newblock Point-slam: Dense neural point cloud-based slam.
\newblock In \emph{Proceedings of the IEEE/CVF International Conference on Computer Vision}, 18433--18444.

\bibitem[{Schmuck and Chli(2017)}]{schmuck2017multi}
Schmuck, P. and Chli, M. (2017).
\newblock Multi-uav collaborative monocular slam.
\newblock In \emph{2017 IEEE International Conference on Robotics and Automation (ICRA)}, 3863--3870. IEEE.

\bibitem[{Schmuck and Chli(2019)}]{schmuck2019ccm}
Schmuck, P. and Chli, M. (2019).
\newblock Ccm-slam: Robust and efficient centralized collaborative monocular simultaneous localization and mapping for robotic teams.
\newblock \emph{Journal of Field Robotics}, 36(4), 763--781.

\bibitem[{Sheng et~al.(2024)Sheng, Mao, Yan, and Yang}]{sheng2024review}
Sheng, X., Mao, S., Yan, Y., and Yang, X. (2024).
\newblock Review on slam algorithms for augmented reality.
\newblock \emph{Displays}, 84, 102806.

\bibitem[{Shorinwa et~al.(2025)Shorinwa, Sun, Schwager, and Majumdar}]{shorinwa2025siren}
Shorinwa, O., Sun, J., Schwager, M., and Majumdar, A. (2025).
\newblock Siren: Semantic, initialization-free registration of multi-robot gaussian splatting maps.
\newblock \emph{arXiv preprint arXiv:2502.06519}.

\bibitem[{Straub et~al.(2019)Straub, Whelan, Ma, Chen, Wijmans, Green, Engel, Mur-Artal, Ren, Verma et~al.}]{straub2019replica}
Straub, J., Whelan, T., Ma, L., Chen, Y., Wijmans, E., Green, S., Engel, J.J., Mur-Artal, R., Ren, C., Verma, S., et~al. (2019).
\newblock The replica dataset: A digital replica of indoor spaces.
\newblock \emph{arXiv preprint arXiv:1906.05797}.

\bibitem[{Sturm et~al.(2012)Sturm, Engelhard, Endres, Burgard, and Cremers}]{sturm2012benchmark}
Sturm, J., Engelhard, N., Endres, F., Burgard, W., and Cremers, D. (2012).
\newblock A benchmark for the evaluation of rgb-d slam systems.
\newblock In \emph{2012 IEEE/RSJ international conference on intelligent robots and systems}, 573--580. IEEE.

\bibitem[{Szot et~al.(2021)Szot, Clegg, Undersander, Wijmans, Zhao, Turner, Maestre, Mukadam, Chaplot, Maksymets, Gokaslan, Vondrus, Dharur, Meier, Galuba, Chang, Kira, Koltun, Malik, Savva, and Batra}]{szot2021habitat}
Szot, A., Clegg, A., Undersander, E., Wijmans, E., Zhao, Y., Turner, J., Maestre, N., Mukadam, M., Chaplot, D., Maksymets, O., Gokaslan, A., Vondrus, V., Dharur, S., Meier, F., Galuba, W., Chang, A., Kira, Z., Koltun, V., Malik, J., Savva, M., and Batra, D. (2021).
\newblock Habitat 2.0: Training home assistants to rearrange their habitat.
\newblock In \emph{Advances in Neural Information Processing Systems (NeurIPS)}.

\bibitem[{Thomas et~al.(2025)Thomas, Sonawalla, Rose, and How}]{thomas2025grand}
Thomas, A., Sonawalla, A., Rose, A., and How, J.P. (2025).
\newblock Grand-slam: Local optimization for globally consistent large-scale multi-agent gaussian slam.
\newblock \emph{arXiv preprint arXiv:2506.18885}.

\bibitem[{Tian et~al.(2022)Tian, Chang, Arias, Nieto-Granda, How, and Carlone}]{tian2022kimera}
Tian, Y., Chang, Y., Arias, F.H., Nieto-Granda, C., How, J.P., and Carlone, L. (2022).
\newblock Kimera-multi: Robust, distributed, dense metric-semantic slam for multi-robot systems.
\newblock \emph{IEEE Transactions on Robotics}, 38(4).

\bibitem[{Tosi et~al.(2024)Tosi, Zhang, Gong, Sandstr{\"o}m, Mattoccia, Oswald, and Poggi}]{tosi2024nerfs}
Tosi, F., Zhang, Y., Gong, Z., Sandstr{\"o}m, E., Mattoccia, S., Oswald, M.R., and Poggi, M. (2024).
\newblock How nerfs and 3d gaussian splatting are reshaping slam: a survey.
\newblock \emph{arXiv preprint arXiv:2402.13255}, 4, 1.

\bibitem[{Wang et~al.(2025)Wang, Guo, Chen, and Lu}]{wang2025depth}
Wang, K., Guo, J., Chen, K., and Lu, J. (2025).
\newblock An in-depth examination of slam methods: Challenges, advancements, and applications in complex scenes for autonomous driving.
\newblock \emph{IEEE Transactions on Intelligent Transportation Systems}.

\bibitem[{Wang et~al.(2004)Wang, Bovik, Sheikh, and Simoncelli}]{wang2004image}
Wang, Z., Bovik, A.C., Sheikh, H.R., and Simoncelli, E.P. (2004).
\newblock Image quality assessment: from error visibility to structural similarity.
\newblock \emph{IEEE transactions on image processing}, 13(4), 600--612.

\bibitem[{Wu et~al.(2024)Wu, Yuan, Zhang, Yang, Cao, Yan, and Gao}]{wu2024recent}
Wu, T., Yuan, Y.J., Zhang, L.X., Yang, J., Cao, Y.P., Yan, L.Q., and Gao, L. (2024).
\newblock Recent advances in 3d gaussian splatting.
\newblock \emph{Computational Visual Media}, 10(4), 613--642.

\bibitem[{Xu et~al.(2025)Xu, Xue, Zhao, Pan, Scherer, and Huang}]{xu2025mac}
Xu, X., Xue, F., Zhao, S., Pan, Y., Scherer, S., and Huang, X. (2025).
\newblock Mac-ego3d: Multi-agent gaussian consensus for real-time collaborative ego-motion and photorealistic 3d reconstruction.
\newblock In \emph{Proceedings of the Computer Vision and Pattern Recognition Conference}, 854--863.

\bibitem[{Yan et~al.(2024)Yan, Qu, Xu, Zhao, Wang, Wang, and Li}]{yan2024gs}
Yan, C., Qu, D., Xu, D., Zhao, B., Wang, Z., Wang, D., and Li, X. (2024).
\newblock Gs-slam: Dense visual slam with 3d gaussian splatting.
\newblock In \emph{Proceedings of the IEEE/CVF Conference on Computer Vision and Pattern Recognition}, 19595--19604.

\bibitem[{Yu et~al.(2025)Yu, Chen, and Schwager}]{yu2025hammer}
Yu, J., Chen, T., and Schwager, M. (2025).
\newblock Hammer: Heterogeneous, multi-robot semantic gaussian splatting.
\newblock \emph{IEEE Robotics and Automation Letters}.

\bibitem[{Yugay et~al.(2025)Yugay, Gevers, and Oswald}]{yugay2025magic}
Yugay, V., Gevers, T., and Oswald, M.R. (2025).
\newblock Magic-slam: Multi-agent gaussian globally consistent slam.
\newblock In \emph{Proceedings of the Computer Vision and Pattern Recognition Conference}, 6741--6750.

\bibitem[{Zhang et~al.(2018)Zhang, Isola, Efros, Shechtman, and Wang}]{zhang2018unreasonable}
Zhang, R., Isola, P., Efros, A.A., Shechtman, E., and Wang, O. (2018).
\newblock The unreasonable effectiveness of deep features as a perceptual metric.
\newblock In \emph{Proceedings of the IEEE conference on computer vision and pattern recognition}, 586--595.

\bibitem[{Zhang et~al.(2022)Zhang, Zhang, Chen, and Zhou}]{zhang2022cvids}
Zhang, T., Zhang, L., Chen, Y., and Zhou, Y. (2022).
\newblock Cvids: A collaborative localization and dense mapping framework for multi-agent based visual-inertial slam.
\newblock \emph{IEEE transactions on image processing}, 31, 6562--6576.

\bibitem[{Zhao et~al.(2024)Zhao, Gao, Wu, Singh, Jiang, Sun, Sarawata, Qiu, Whittaker, Higgins et~al.}]{zhao2024subt}
Zhao, S., Gao, Y., Wu, T., Singh, D., Jiang, R., Sun, H., Sarawata, M., Qiu, Y., Whittaker, W., Higgins, I., et~al. (2024).
\newblock Subt-mrs dataset: Pushing slam towards all-weather environments.
\newblock In \emph{Proceedings of the IEEE/CVF Conference on Computer Vision and Pattern Recognition}, 22647--22657.

\bibitem[{Zhong et~al.(2023)Zhong, Qi, Chen, Wu, Chen, and Liu}]{zhong_dcl-slam_2023}
Zhong, S., Qi, Y., Chen, Z., Wu, J., Chen, H., and Liu, M. (2023).
\newblock Dcl-slam: {A} distributed collaborative lidar slam framework for a robotic swarm.
\newblock \emph{IEEE sensors journal}, 24(4), 4786--4797.
\newblock Publisher: IEEE.

\bibitem[{Zhu et~al.(2025)Zhu, Li, Sandstr{\"o}m, Huang, Schindler, and Armeni}]{zhu2025loopsplat}
Zhu, L., Li, Y., Sandstr{\"o}m, E., Huang, S., Schindler, K., and Armeni, I. (2025).
\newblock Loopsplat: Loop closure by registering 3d gaussian splats.
\newblock In \emph{2025 International Conference on 3D Vision (3DV)}, 156--167. IEEE.

\bibitem[{Zhu et~al.(2024)Zhu, Wang, Kong, Kong, and Wang}]{zhu20243d}
Zhu, S., Wang, G., Kong, X., Kong, D., and Wang, H. (2024).
\newblock 3d gaussian splatting in robotics: A survey.
\newblock \emph{arXiv preprint arXiv:2410.12262}.

\bibitem[{Zhu et~al.(2023)Zhu, Kong, Jie, Xu, and Cheng}]{zhu2023graco}
Zhu, Y., Kong, Y., Jie, Y., Xu, S., and Cheng, H. (2023).
\newblock Graco: A multimodal dataset for ground and aerial cooperative localization and mapping.
\newblock \emph{IEEE Robotics and Automation Letters}, 8(2), 966--973.

\bibitem[{Zobeidi et~al.(2022)Zobeidi, Koppel, and Atanasov}]{zobeidi2022dense}
Zobeidi, E., Koppel, A., and Atanasov, N. (2022).
\newblock Dense incremental metric-semantic mapping for multiagent systems via sparse gaussian process regression.
\newblock \emph{IEEE Transactions on Robotics}, 38(5), 3133--3153.

\bibitem[{Zou and Tan(2013)}]{zou2012coslam}
Zou, D. and Tan, P. (2013).
\newblock Coslam: Collaborative visual slam in dynamic environments.
\newblock \emph{IEEE transactions on pattern analysis and machine intelligence}, 35(2), 354--366.

\bibitem[{Zou et~al.(2019)Zou, Tan, and Yu}]{zou2019collaborative}
Zou, D., Tan, P., and Yu, W. (2019).
\newblock Collaborative visual slam for multiple agents: A brief survey.
\newblock \emph{Virtual Reality \& Intelligent Hardware}, 1(5), 461--482.

\end{thebibliography}
                                                   







\end{document}